\documentclass[journal]{IEEEtran}

\usepackage{xcolor,soul,framed} 

\colorlet{shadecolor}{yellow}
\usepackage[pdftex]{graphicx}
\graphicspath{{../pdf/}{../jpeg/}}
\DeclareGraphicsExtensions{.pdf,.jpeg,.png}

\ifCLASSOPTIONcompsoc
  \usepackage[caption=false,font=normalsize,labelfont=sf,textfont=sf]{subfig}
\else
  \usepackage[caption=false,font=footnotesize]{subfig}
\fi

\usepackage[cmex10]{amsmath}
\usepackage{bm}
\usepackage{array}
\usepackage{tabularx}
\usepackage{mdwmath}
\usepackage{mdwtab}
\usepackage{amsfonts}
\usepackage{amssymb}
\usepackage{amsthm}
\usepackage{eqparbox}
\usepackage{url}
\usepackage{mathtools}
\usepackage{tensor}
\usepackage{multirow}
\usepackage{tikz}
\usetikzlibrary{chains, fit, shapes, decorations.pathreplacing, decorations.pathmorphing, shapes.symbols, calc, trees, positioning, arrows, arrows.meta, shapes.geometric, matrix}
\usepackage{enumerate}
\usepackage{hyperref}
\usepackage{algorithm}
\usepackage{algorithmic}
\usepackage{cite}

\allowdisplaybreaks
\newtheorem{assumption}{Assumption}
\newtheorem{definition}{Definition}

\begin{document}
\title{Multi-objective Anti-swing Trajectory Planning of Double-pendulum Tower Crane Operations using Opposition-based Evolutionary Algorithm}
\author{
Souravik Dutta,~%
Yiyu Cai,~%
and~Jianmin Zheng~%
\thanks{This work was funded in part by the Energy Research Institute @NTU (ERI@N) of Nanyang Technological University (NTU), Singapore, through National Research Foundation fellowship. (\textit{Corresponding author: Yiyu Cai.})}
\thanks{S. Dutta is with School of Mechanical and Aerospace Engineering (MAE) and Energy Research Institute @NTU (ERI@N) - Interdisciplinary Graduate Programme (IGP) of Nanyang Technological University, Singapore 639798 (e-mail: SOURAVIK001@e.ntu.edu.sg).}
\thanks{Y. Cai is with School of Mechanical and Aerospace Engineering (MAE), and Energy Research Institute @NTU (ERI@N) of Nanyang Technological University, Singapore 639798 (e-mail: MYYCai@ntu.edu.sg).}
\thanks{J. Zheng is with School of Computer Science and Engineering (CSE) of Nanyang Technological University, Singapore 639798 (e-mail: ASJMZheng@ntu.edu.sg).}
}


\maketitle

\begin{abstract}

Underactuated tower crane lifting requires time-energy optimal trajectories for the trolley/slew operations and reduction of the unactuated swings resulting from the trolley/jib motion. In scenarios involving non-negligible hook mass or long rig-cable, the hook-payload unit exhibits double-pendulum behaviour, making the problem highly challenging. This article introduces an offline multi-objective anti-swing trajectory planning module for a Computer-Aided Lift Planning (CALP) system of autonomous double-pendulum tower cranes, addressing all the transient state constraints. A set of auxiliary outputs are selected by methodically analyzing the payload swing dynamics and are used to prove the differential flatness property of the crane operations. The flat outputs are parameterized via suitable Bézier curves to formulate the multi-objective trajectory optimization problems in the flat output space. A novel multi-objective evolutionary algorithm called Collective Oppositional Generalized Differential Evolution 3 (CO-GDE3) is employed as the optimizer. To obtain faster convergence and better consistency in getting a wide range of good solutions, a new population initialization strategy is integrated into the conventional GDE3. The computationally efficient initialization method incorporates various concepts of computational opposition. Statistical comparisons based on trolley and slew operations verify the superiority of convergence and reliability of CO-GDE3 over the standard GDE3. Trolley and slew operations of a collision-free lifting path computed via the path planner of the CALP system are selected for a simulation study. The simulated trajectories demonstrate that the proposed planner can produce time-energy optimal solutions, keeping all the state variables within their respective limits and restricting the hook and payload swings.

\end{abstract}


\IEEEpeerreviewmaketitle


\section{Introduction}
\label{sec:introduction}

\IEEEPARstart{T}{ower} cranes act as underactuated systems, producing nonlinear coupling between actuated trolley/jib motion and unactuated swings of the hoist-cable and payload unit. A feedforward approach to control the crane motion during autonomous lifting is to optimally design trajectories of the actuated operations, which can effectively restrict unwanted swings. The solution to the trajectory problem generates necessary reference inputs for the tracking control system of the tower crane. Hence, an anti-swing trajectory can enable any conventional controller to track the desired motion with reduced spherical pendulum behaviour during and at the end of the operation, without the requirement of a separate anti-swing controller. Most of the studies in the literature address this scenario as a single-pendulum problem \cite{Sun2016,Le2017,Sun2019,Chen2019,Liu2019,Zhang2020}. However, two pendulum motions exist if the mass of the hook is comparable to that of the payload or the length of the rig-cable connecting the hook and the payload is significant because of the size of the payload. The hoist-cable and the hook constitute the first spherical pendulum system, and the second one consists of the rig-cable and the payload. To follow a planned collision-free lifting path, the swing amplitudes should be as low as possible and later eliminated at the end of an operation, to maintain the safety of the planned path. To ensure sufficient productivity and safety of the actuators, the designed trajectory of a lifting path should be optimal and jerk-continuous. A smooth optimal trajectory with motion-induced swing suppression and residual swing reduction can provide a safe and productive control action.

Evolutionary Algorithms (EAs) have been extensively utilized for solving path and trajectory planning problems of robotics and autonomous systems \cite{Wu2021}. The literature spans several application areas, including robotic manipulators \cite{Sadhu2018}, mobile robots\cite{Das2016}, unmanned aerial vehicles \cite{Chen2022,Lin2022}, and single-pendulum cranes \cite{Liu2021}. This applicability is especially due to the intrinsic optimization requirements of planning problems and the inherent ability of EAs to generate a population of solutions based on the trade-off between the multiple objectives that are pertinent in robotic planning exercises. Therefore, EAs have a strong potential in obtaining time-energy optimal solutions for the trajectory planning problem of autonomous double-pendulum tower cranes.

\subsection{Related work in literature}
\label{subsec:litreview}

To achieve the objectives of trolley/jib positioning and hook-payload swing restriction at the same time, researchers have incorporated several approaches. Over the last decade, the underactuated double-pendulum characteristics have been investigated majorly for overhead cranes. These approaches include parallel distributed fuzzy linear-quadratic regulator (LQR) control \cite{Adeli2011}, conventional and hierarchical sliding mode control \cite{Tuan2013}, linear matrix inequality (LMI) based robust state feedback control \cite{Ouyang2018}, energy-based nonlinear coupling control \cite{Shi2019}, hybrid command-shaper \cite{Masoud2012}, and optimal trajectory planning \cite{Chen2017,Zhang2021}. For tower cranes, the problem becomes more complex due to the additional complicated slew motion of the jib, which produces spherical movements of the double-pendulum. Thus, high nonlinearity stems from the enhanced coupling between all the degrees of freedom (DOFs), actuated and unactuated.

Recently, numerous studies have been conducted on the double-swing rejection of tower cranes. These can be categorized according to energy-based methods, adaptive tracking control methods, and motion planning methods. Zhang et al. \cite{Zhang2019} have developed a controller based on the total energy of the tower crane system with double-pendulum and spherical-pendulum effects. Lyapunov techniques and LaSalle’s invariance theorem are employed to demonstrate the closed-loop system's asymptotic stability and the system states' convergence. In the work by Ouyang et al. \cite{Ouyang2020c}, a nonlinear controller based on energy-shaping is presented to achieve both slew/translation positioning and suppression of the double-pendulum sway angles. A nonlinear adaptive tracking controller is designed in \cite{Ouyang2020} to solve the above problems without a linearized crane model. The proposed controller estimates the model's unknown parameters, demonstrating sufficient control performance under different model parameters. Zhang et al. \cite{Zhang2021a} have systematically investigated a novel adaptive neural network tracking control method for a unique double-pendulum tower crane system model. Neural networks are employed to approximate the functions with uncertain/unknown dynamics and non-ideal inputs. A partial enhanced-coupling nonlinear controller with initial saturation is designed in \cite{Tian2021}. Lyapunov technique and LaSalle’s invariance theorem are used to analyze the controller's stability. In \cite{Ouyang2021}, rapid and effective control performance is reported by enhancing the coupling between the actuated and the unactuated parts. Ouyang et al. \cite{Ouyang2020a} have designed a composite trajectory planning strategy for tower cranes with a double-pendulum effect. The first part of the proposed trajectory ensures the positioning of the jib and trolley, and the second part secures swing suppression by using a damping component. The only optimal motion planning approach for the double-swing reduction in tower cranes is catalogued by Li et al. \cite{Li2022}, who have considered auxiliary signals to express the state variables and introduced time-based polynomial functions for trajectory planning.

Apart from the methods by \cite{Ouyang2020a} and \cite{Li2022}, there is no trajectory solution to the issue of double-pendulum swing suppression of tower cranes. The existing closed-loop control solutions work with only feasible reference trajectories while achieving the effective payload swing reduction induced by the crane operations via feedback control. These methods usually do not address the state constraints arising due to mechanical limits and safety concerns. Moreover, there is \textit{no published work} for time-energy optimal trajectory generation of double-pendulum tower cranes, which is pertinent in spending optimal effort to gain optimal output. This is particularly due to the highly nonlinear nature of the slew motion, which leads to coupling between all the state variables. As a result, parameterization of the state DOFs becomes a challenging case, raising difficulty in the subsequent formulation of the trajectory optimization problems. Hence, there is a need for a multi-objective trajectory planning method for autonomous double-pendulum tower cranes, which can minimize the operating time as well as the operating energy, and constrain the swing of the hook and the payload during lifting.

Multi-Objective EAs (MOEAs) improve (evolve) a population of solutions through a finite number of iterations (generations), via incremental perturbations, to find a set of known Pareto optimal solutions \cite{Miettinen1998} to multi-objective problems such as the task of time-energy optimal trajectory planning. These solutions form the known Pareto front (called $\mathcal{PF}_{known}$) in the objective space. However, the quality of the Pareto optimal solutions generated by typical MOEAs and the algorithms' convergence speed are sensitive to the initial population.  The default approach, in the absence of \textit{a priori} knowledge about the solution landscape, is to initialize the population of solutions randomly sampled with uniform probability distribution within the decision interval. By improving the initialization method, the algorithm performance can be improved in terms of faster convergence to $\mathcal{PF}_{known}$ and a diverse set of better values throughout the objective space. In EA literature, several population initialization techniques have been employed in evolutionary algorithms (EAs) for single-objective problems \cite{Kazimipour2014}. For MOEAs, very few studies have proposed unconventional population initialization methods, such as seeding using gradient-based information \cite{HernandezDiaz2008}, evolutionary strategy (ES) \cite{Friedrich2015}, hybrid chaotic mapping model \cite{Liu2015} and opposition-based learning (OBL) \cite{Bhowmik2015,Wang2019,Ewees2021}. Preliminary local search (like the gradient-based method proposed in \cite{HernandezDiaz2008}) or machine learning techniques incur a certain computational budget. The hybrid chaotic mapping model of \cite{Liu2015} considers an extra control parameter just to generate the initial population. In the opposition-based techniques presented in \cite{Friedrich2015,Bhowmik2015,Ewees2021}, either scalar or vector fitness-based evaluations are conducted at the initialization stage to generate better solutions, for quicker convergence to $\mathcal{PF}_{known}$. The approaches employing domination-based selection also require a pruning step to obtain a population of the required size, which again adds to the computational cost of the algorithm. So, there is a huge scope for improvement in the context of new computationally efficient population initialization techniques for standard MOEAs, and the application of the same for multi-objective anti-swing trajectory planning of autonomous double-pendulum tower crane operations.

\subsection{Contributions of proposed work}
\label{subsec:researchconstributions}

According to the best of the authors' knowledge, the current work proposes the \textit{first} offline multi-objective anti-swing trajectory planner for autonomous double-pendulum tower cranes, that can plan time-energy optimal trajectories while addressing all the transient constraints imposed on the operations. This planning module is designed as a part of the Computer-Aided Lift Planning (CALP) System \cite{Cai2016,Dutta2020}, to take collision-free paths as input from the path planner and provide optimal trajectories as output. The contributions of the proposed work are twofold:

\begin{enumerate}
    \item The trajectory planner is constructed to deal with the complex dynamics of the double-swing behaviour of the hook-payload unit during trolley/jib positioning by exploiting the differential flatness of the tower crane system during the trolley and slew motions. Subsequently, parameterization of the flat outputs by suitable B\'{e}zier curves is conducted to transform the trajectory planning tasks of the crane operations into constrained multi-objective trajectory optimization problems (MOTOPs).
    
    \item To solve the formulated optimization problems, Generalized Differential Evolution 3 (GDE3) \cite{Kukkonen2005} is integrated with a novel population initialization method, named Collective Oppositional initialization. The objective of the new initialization strategy is to obtain faster convergence and better consistency in getting a wide range of good solutions, without adding to the computational complexity of the conventional algorithm. An average fuzzy membership function is utilized to evaluate the output solutions of the designed optimizer and decide the best operating time and operating effort for the problem in question.
\end{enumerate} 

Simulation studies involving both trolley and slew operations are performed to statistically compare the improved MOEA with its conventional counterpart and to validate the effectiveness and efficiency of the proposed optimal trajectory planner.

\section{Differential Flatness of Double-pendulum Tower Crane Operations}
\label{sec:towercranemodel}

According to Fliess et al. \cite{Fliess1992}, a system is differentially flat if there exists an output or there exist several outputs, such that the state and input variables can be parameterized by the output(s) and a finite number of its(their) time derivatives. Such auxiliary outputs are called the flat outputs of the system. If the differential flatness of the tower crane system can be proven, the trajectory planning for the operations can be done in the flat output space, as all the DOFs (and hence their time derivatives) are represented by the flat output(s) and its(their) time derivatives.

\begin{figure}[!t]
    \begin{center}
        \includegraphics[width=\columnwidth]{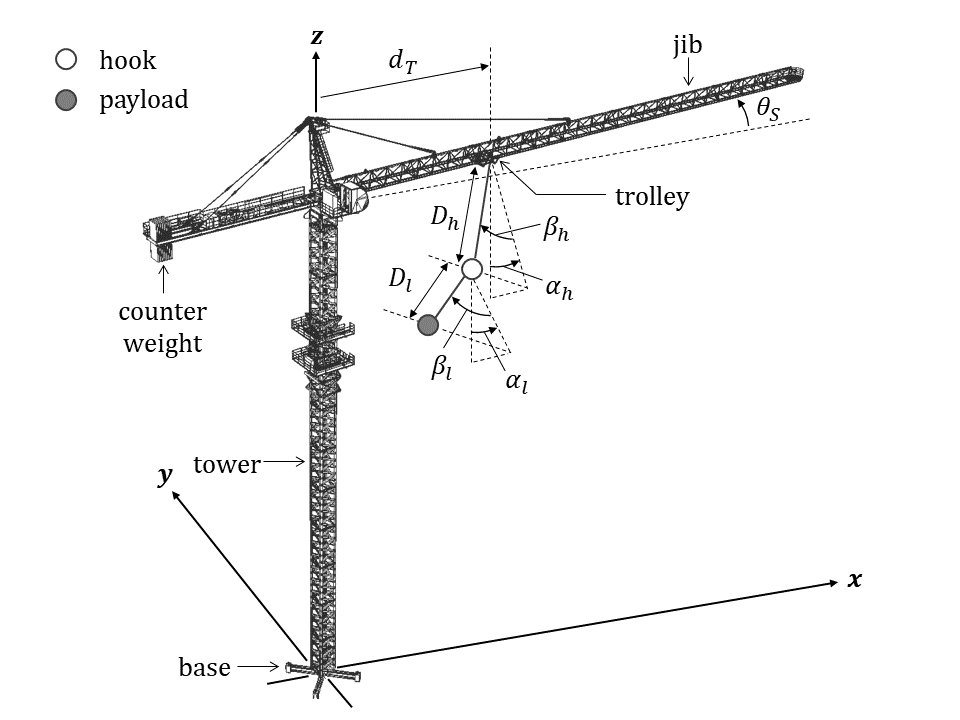}
        \caption{Structure of double-pendulum tower crane with actuated and unactuated DOFs.}
        \label{fig:craneDOFs}
    \end{center}
\end{figure}

Since the focus of the present research is to find optimal trajectories to suppress the payload swings, only the trolley and slew motions are considered. Referring to Fig. \ref{fig:craneDOFs}, the tower crane is represented as an underactuated system with 6 DOFs. The DOFs representing the fundamental crane operations are slew angle $\theta_S$ and trolley distance $d_T$ along the jib of the crane. The fixed length of the hoist-cable and the fixed distance of the rig-cable between the hook and the payload are denoted by $D_h$ and $D_l$, respectively. The radial swing angles of the hook and the payload are denoted by $\alpha_h$ and $\alpha_l$, while the tangential swing angles of the hook and the payload are defined as $\beta_h$ and $\beta_l$, respectively.

Without loss of generality, the following assumptions regarding tower crane lifting are made to develop the robotized model of the crane:

\begin{assumption}
    The massless hoist-cable and rig-cable do not undergo twisting or stretching under load. The hook and the payload are also modelled as point masses. The hoist-cable, the hook, the rig-cable, and the payload, together constitute the spherical double-pendulum system.
\end{assumption}

\begin{assumption}
    In the course of the crane operations, the hook and the payload are always beneath the jib-trolley section \cite{Sun2012a,Sun2013a,Huang2013}. Mathematically:
    
    \begin{equation}
        \alpha_h,\beta_h, \alpha_l, \beta_l \in \left(-\frac{\pi}{2},\frac{\pi}{2}\right)
        \label{eq:swinglimits}
    \end{equation}
\end{assumption}

\begin{assumption}
    Given the payload swing has to be kept as low as possible, the small-angle approximation holds for all swing angles ($\alpha_h,\beta_h, \alpha_l, \beta_l < 5^\circ$) during the operation \cite{Sun2013}. This implies:
    
    \begin{equation}
        \begin{aligned}
            & \sin{\phi} \approx \phi, \; \cos{\phi} \approx 1, \; \phi^2 \approx \dot{\phi}^2 \approx 0, \; \phi\psi \approx \dot{\phi}\dot{\psi} \approx 0; \\
            & \forall \phi \neq \psi; \; \phi, \psi \in \left\{ \alpha_h, \beta_h, \alpha_l, \alpha_l \right\}
        \end{aligned}
        \label{eq:smallangle}
    \end{equation}
\end{assumption}

\begin{assumption}
    Frictional effects on the actuated motions and air resistance on the hook-payload unit are negligible. No strong wind is present during the lifting operations.
\end{assumption}

For convenience, unless stated otherwise, the following convention is used throughout the manuscript:

\begin{equation}
    \cos{\phi} \overset{\Delta}{=} C_{\phi}, \; \sin{\phi} \overset{\Delta}{=} S_{\phi}; \quad \forall \phi \in \left\{ \theta_S, \alpha_h, \beta_h, \alpha_l, \beta_l \right\}
\end{equation}

\subsection{Flat output of double-pendulum trolley operation}
\label{subsec:trolleydynamics}

\begin{figure}[!t]
    \begin{center}   
        \includegraphics[width=\columnwidth]{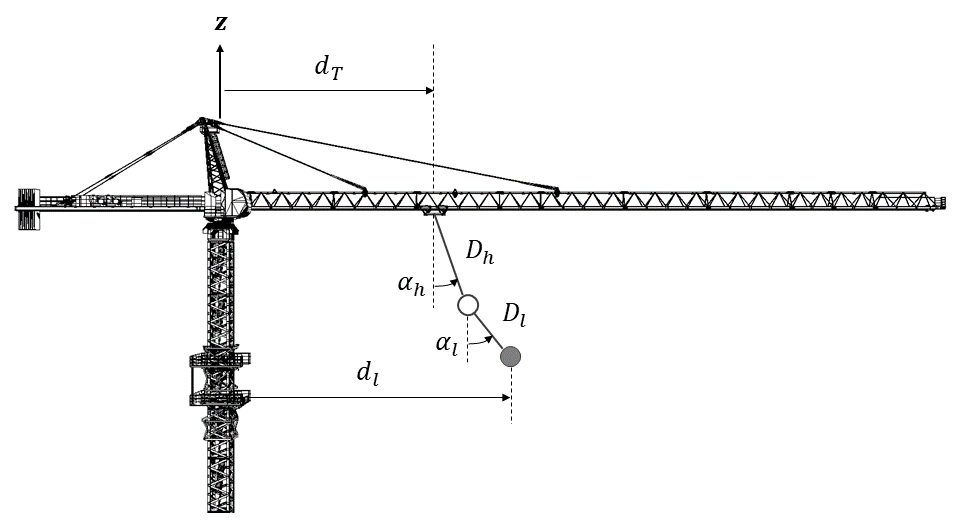}
        \caption{Trolley motion of double-pendulum tower crane with its auxiliary output.}
        \label{fig:trolleyauxDOFs}
    \end{center}
\end{figure}

For pure trolley motion, the system behaviour is illustrated in Fig. \ref{fig:trolleyauxDOFs}. Only the radial swing angles $\alpha_h$ and $\alpha_l$ are excited due to the actuated movement, exhibiting planar double-pendulum responses. Let $D_H = D_h+D_l$ be the effective hoist height during the trolley transportation, and the combined mass of the hook and the trolley be denoted by $m_{hl} = m_h+m_l$. Implementing Eq. (\ref{eq:smallangle}), the corresponding unactuated hook-payload dynamics can be described as:

\begin{equation}
    \ddot{d_T} + D_h\ddot{\alpha_h} + \frac{m_l}{m_{hl}}D_l\ddot{\alpha_l} + g\alpha_h = 0
    \label{eq:trolleyhsimp}
\end{equation}

\begin{equation}
    \ddot{d_T} + D_h\ddot{\alpha_h} + D_l\ddot{\alpha_l} + g\alpha_l = 0
    \label{eq:trolleylsimp}
\end{equation}

The above set of equations establishes the coupling relationship among the swing angle outputs of the hook and the payload, and the trolley acceleration, which acts as the input. Apart from these state and input variables, the position of the payload measured along the jib, $d_l$, can be considered as an auxiliary output, as shown in Fig. \ref{fig:trolleyauxDOFs}.

Through geometric analysis of the trolley-hook-payload motion, and applying the small-angle approximation of Eq. (\ref{eq:smallangle}), it is computed that:

\begin{equation}
    d_l = d_T + D_h\alpha_h + D_l\alpha_l
    \label{eq:trolleyauxDOF}
\end{equation}

Differentiating Eq. (\ref{eq:trolleyauxDOF}) twice with respect to time:

\begin{equation}
    \ddot{d_l} = \ddot{d_T} + D_h\ddot{\alpha_h} + D_l\ddot{\alpha_l}
    \label{eq:trolleyauxDOFacc}
\end{equation}

Now, substituting Eq. (\ref{eq:trolleyauxDOFacc}) in Eqs. (\ref{eq:trolleyhsimp} - \ref{eq:trolleylsimp}), it is deduced that:

\begin{equation}
    \ddot{d_l} - \frac{m_h}{m_{hl}}D_l\ddot{\alpha_l} + g\alpha_h = 0
    \label{eq:trolleyalphahfpsimp}
\end{equation}

\begin{equation}
    \ddot{d_l} + g\alpha_l = 0
    \label{eq:trolleylfpsimp}
\end{equation}

Rearranging, Eq. (\ref{eq:trolleylfpsimp}) can be written as:

\begin{equation}
    \alpha_l = -\frac{1}{g}\ddot{d_l}
    \label{eq:trolleylfp}
\end{equation}

Substituting Eq. (\ref{eq:trolleylfp}) in Eq. (\ref{eq:trolleyalphahfpsimp}):

\begin{equation}
    \alpha_h = -\frac{1}{g}\ddot{d_l} + \frac{m_hD_l}{m_{hl}g^2}{d_l}^{(4)}
    \label{eq:trolleyhfp}
\end{equation}

Then, substituting $\alpha_h$ and $\alpha_l$ in Eq. (\ref{eq:trolleyauxDOF}), the following is obtained:

\begin{equation}
    d_T = d_l + \frac{D_H}{g}\ddot{d_l} - \frac{m_hD_hD_l}{m_{hl}g^2}{d_l}^{(4)}
    \label{eq:trolleydTfp}
\end{equation}

Based on Eqs. (\ref{eq:trolleylfp} - \ref{eq:trolleydTfp}), the actuated DOF $d_T$, and the unactuated DOFs $\alpha_h$ and $\alpha_l$, can be represented by the auxiliary DOF $d_l$, and some of its time derivatives. Hence, it can be concluded that $d_l$ is the flat output of the trolley-hook-payload system. Differentiating Eq. (\ref{eq:trolleydTfp}) twice with respect to time, the other state variables during the trolley motion are represented as the following:

\begin{equation}
    \dot{d_T} = \dot{d_l} + \frac{D_H}{g}{d_l}^{(3)} - \frac{m_hD_hD_l}{m_{hl}g^2}{d_l}^{(5)}
    \label{eq:trolleydTvelfp}
\end{equation}

\begin{equation}
    \ddot{d_T} = \ddot{d_l} + \frac{D_H}{g}{d_l}^{(4)} - \frac{m_hD_hD_l}{m_{hl}g^2}{d_l}^{(6)}
    \label{eq:trolleydTaccfp}
\end{equation}

\subsection{Flat outputs of double-pendulum slew operation}
\label{subsec:slewdynamics}

\begin{figure}[!t]
    \begin{center}   
        \includegraphics[width=\columnwidth]{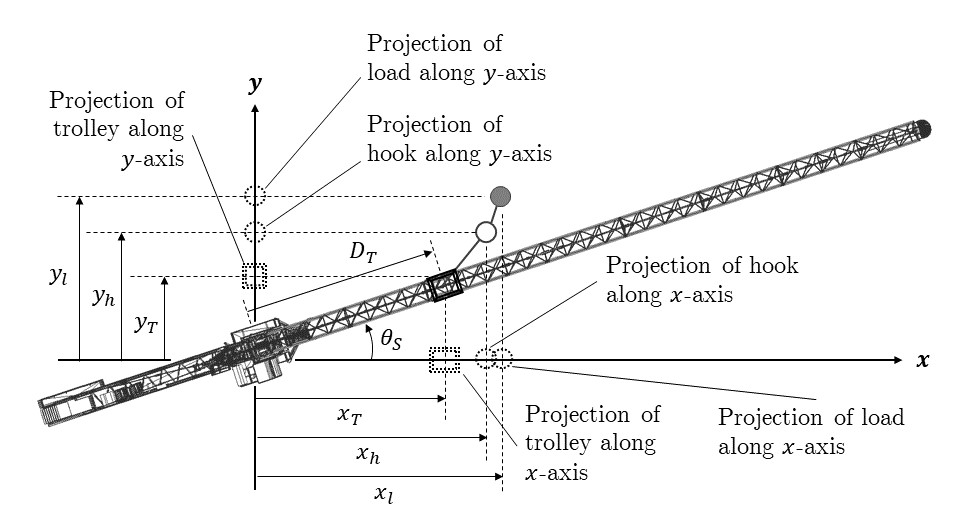}
        \caption{Slew motion of double-pendulum tower crane with its auxiliary outputs.}
        \label{fig:slewdynamicstop}
    \end{center}
\end{figure}

For pure slew motion, the system demonstrates more complex behaviour, as can be interpreted from Fig. \ref{fig:slewdynamicstop}. The inertial and centrifugal forces of the slewing action of the crane jib generate both radial ($\alpha_h$ for hook and $\alpha_l$ for payload) and tangential ($\beta_h$ for hook and $\beta_l$ for payload) swing components, as shown in Fig. \ref{fig:craneDOFs}. Consequently, the hook-payload combination undergoes a spherical double-pendulum motion. From Fig. \ref{fig:slewdynamicstop}, $D_T$ is the fixed distance of the trolley along the jib while slewing. The auxiliary outputs considered are $x_h$, $y_h$, and $x_l$, $y_l$, which are the positions of the hook and the payload, respectively, along the $x$ and the $y$ axis. Analyzing the geometry of the jib-hook-payload motion:

\begin{equation}
    x_h = D_TC_{\theta_S} + D_h\alpha_hC_{\theta_S} - D_h\beta_hS_{\theta_S}
    \label{eq:slewauxDOFxh}
\end{equation}

\begin{equation}
    y_h = D_TS_{\theta_S} + D_h\alpha_hS_{\theta_S} + D_h\beta_hC_{\theta_S}
    \label{eq:slewauxDOFyh}
\end{equation}

\begin{equation}
    x_l = x_h + D_l\alpha_lC_{\theta_S} - D_l\beta_lS_{\theta_S}
    \label{eq:slewauxDOFxl}
\end{equation}

\begin{equation}
    y_l = y_h + D_l\alpha_lS_{\theta_S} + D_l\beta_lC_{\theta_S}
    \label{eq:slewauxDOFyl}
\end{equation}

Applying basic mathematical operations on the expressions of $x_h$, $y_h$, and $x_l$, $y_l$, it is easily computed that:

\begin{equation}
    \alpha_h = \frac{1}{D_h} \left( x_hC_{\theta_S} + y_hS_{\theta_S} - D_T \right)
    \label{eq:slewahthetas}
\end{equation}

\begin{equation}
    \beta_h = \frac{1}{D_h} \left( y_hC_{\theta_S} - x_hS_{\theta_S} \right)
    \label{eq:slewbhthetas}
\end{equation}

\begin{equation}
    \alpha_l = \frac{1}{D_l} \left( x_lC_{\theta_S} + y_lS_{\theta_S} - x_hC_{\theta_S} - y_hS_{\theta_S} \right)
    \label{eq:slewalthetas}
\end{equation}

\begin{equation}
    \beta_l = \frac{1}{D_l} \left( y_lC_{\theta_S} - x_lS_{\theta_S} + x_hS_{\theta_S} - y_hC_{\theta_S} \right)
    \label{eq:slewblthetas}
\end{equation}

While the jib rotates on the $xy$ plane about the $z$ axis of the global coordinate system, two simultaneous trolley displacements, $x_T$ and $y_T$, occur in two perpendicular directions due to the pure slewing motion of the tower crane. Exploiting this fact, the slew operation is decomposed into two orthogonally projected trolley operations along the $x$ and the $y$ axes of the fixed reference system. For a trolley motion along a fixed direction, the trolley position, and positions of the hook and the payload along that direction, are related according to Eq. (\ref{eq:trolleydTfp}), as derived in Section \ref{subsec:trolleydynamics}. Utilizing this association, the following equation can be deduced for the projection of trolley motion on the $x$ axis:

\begin{equation}
    x_T = x_l + \frac{D_H}{g}\ddot{x_l} - \frac{m_hD_hD_l}{m_{hl}g^2}{x_l}^{(4)}
    \label{eq:slewxTfp}
\end{equation}


Based on Fig. \ref{fig:slewdynamicstop}, the following relationship is calculated from Eq. (\ref{eq:slewxTfp}):

\begin{equation}
    C_{\theta_S} = \frac{x_T}{D_T} = \frac{1}{D_T}x_l + \frac{D_H}{gD_T}\ddot{x_l} - \frac{m_hD_hD_l}{m_{hl}D_Tg^2}{x_l}^{(4)}
    \label{eq:slewthetaSfpx}
\end{equation}

Therefore, the slew angle can be expressed in terms of the auxiliary DOFs during the slew motion as:

\begin{equation}
    \begin{aligned}
        \theta_S & = C^{-1}\left( \frac{x_T}{D_T} \right) \\
        & = C^{-1}\left( \frac{1}{D_T}x_l + \frac{D_H}{gD_T}\ddot{x_l} - \frac{m_hD_hD_l}{m_{hl}D_Tg^2}{x_l}^{(4)} \right)
    \end{aligned}
    \label{eq:slewthetaSfp}
\end{equation}

Here, $C^{-1}(A) = \arccos{(A)}$. Now, substituting Eqs. (\ref{eq:slewthetaSfpx} - \ref{eq:slewthetaSfp}) in Eqs. (\ref{eq:slewahthetas} - \ref{eq:slewblthetas}), the following expressions of the swing angles are calculated:

\begin{equation}
    \alpha_h = \frac{1}{D_h} \Bigg[ x_h\left( \frac{x_T}{D_T} \right) + y_h\sin{\left[ C^{-1}\left( \frac{x_T}{D_T} \right) \right]} - D_T \Bigg]
    \label{eq:slewahfp}
\end{equation}

\begin{equation}
    \beta_h = \frac{1}{D_h} \Bigg[ y_h\left( \frac{x_T}{D_T} \right) - x_h\sin{\left[ C^{-1}\left( \frac{x_T}{D_T} \right) \right]} \Bigg]
    \label{eq:slewbhfp}
\end{equation}

\begin{equation}
    \begin{aligned}
        \alpha_l = & \frac{1}{D_l} \Bigg[ x_l\left( \frac{x_T}{D_T} \right) + y_L\sin{\left[ C^{-1}\left( \frac{x_T}{D_T} \right) \right]} \\
        & - x_h\left( \frac{x_T}{D_T} \right) - y_h\sin{\left[ C^{-1}\left( \frac{x_T}{D_T} \right) \right]} \Bigg]
    \end{aligned}
    \label{eq:slewalfp}
\end{equation}

\begin{equation}
    \begin{aligned}
        \beta_l = & \frac{1}{D_l} \Bigg[ y_l\left( \frac{x_T}{D_T} \right) - x_l\sin{\left[ C^{-1}\left( \frac{x_T}{D_T} \right) \right]} \\
        & + x_h\sin{\left[ C^{-1}\left( \frac{x_T}{D_T} \right) \right]} - y_h\left( \frac{x_T}{D_T} \right) \Bigg]
    \end{aligned}
    \label{eq:slewblfp}
\end{equation}

In the aforementioned equations, $x_T$ is expressed as Eq. (\ref{eq:slewxTfp}). From Eqs. (\ref{eq:slewthetaSfp} - \ref{eq:slewblfp}), it can be inferred that the hook and the payload coordinates $x_h$, $y_h$, $x_l$ and $y_l$ are the flat outputs of the jib-hook-payload system. To represent the other state variables of the slew motion, differentiating Eq. (\ref{eq:slewthetaSfpx}) twice with respect to time, the following can be derived:

\begin{equation}
    \dot{\theta_S} = -\frac{1}{S_{\theta_S}} \left[ \frac{1}{D_T}\dot{x_l} + \frac{D_H}{gD_T}{x_l}^{(3)} - \frac{m_hD_hD_l}{m_{hl}D_Tg^2}{x_l}^{(5)} \right]
    \label{eq:slewthetaSvelfp}
\end{equation}

\begin{equation}
    \begin{aligned}
        \ddot{\theta_S} = & -\frac{C_{\theta_S}}{S_{\theta_S}^3} \left[ \frac{1}{D_T}\dot{x_l} + \frac{D_H}{gD_T}{x_l}^{(3)} - \frac{m_hD_hD_l}{m_{hl}D_Tg^2}{x_l}^{(5)} \right]^2 \\
        & - \frac{1}{S_{\theta_S}} \left[ \frac{1}{D_T}\ddot{x_l} + \frac{D_H}{gD_T}{x_l}^{(4)} - \frac{m_hD_hD_l}{m_{hl}D_Tg^2}{x_l}^{(6)} \right]
    \end{aligned}
    \label{eq:slewthetaSaccfp}
\end{equation}

Hence, the differential flatness of the double-pendulum tower crane system is established for both trolley and slew operation. This indicates that the trajectory planning for the tower crane can be conducted in the flat output space, where all the input and state variables are transformed into their respective flat output expressions.

\section{Trajectory Optimization Problems in Flat Output Space}
\label{sec:motop}

For optimal anti-swing trajectory generation, certain constraints need to be implemented according to the mechanical limits of the crane and the motion safety. Thereafter, in conjunction with the flat output trajectory parameterization, the Multi-Objective Trajectory Optimization Problem (MOTOP) for each operation can be formulated in the respective flat output spaces.

\subsection{Trolley trajectory constraints}
\label{subsec:trolleymotop}

Let $t_T$ be the total trolley transportation time. The velocities and accelerations of the trolley and the hook-payload swings at the initial and final frames of the motion should be zero. The trolley jerk is also required to be zero at the boundaries to ensure jerk-continuity. As the trolley translates, the velocities and accelerations of the trolley and the unactuated swings are to be kept within their respective specified range, to abide by the mechanical limits of the crane and its actuators.

Utilizing the expressions of the trolley distance, the hook swing angle and the payload swing angle from Eqs. (\ref{eq:trolleylfp} - \ref{eq:trolleydTfp}), the aforementioned constraints can be transformed to the corresponding constraints of the trolley flat outputs $d_l$ as:

\begin{equation}
    \begin{aligned}
        & d_l(0) = d_{Ti}, \; \dot{d_l}(0) = \ddot{d_l}(0) = d_l^{(3)}(0) = 0, \\
        & d_l^{(4)}(0) = d_l^{(5)}(0) = d_l^{(6)}(0) = d_l^{(7)}(0) = 0
    \end{aligned}
    \label{eq:trolleyfpeqconstinitial}
\end{equation}

\begin{equation}
    \begin{aligned}
        & d_l(t_T) = d_{Tf}, \; \dot{d_l}(t_T) = \ddot{d_l}(t_T) = d_l^{(3)}(t_T) = 0, \\
        & d_l^{(4)}(t_T) = d_l^{(5)}(t_T) = d_l^{(6)}(t_T) = d_l^{(7)}(t_T) = 0
    \end{aligned}
    \label{eq:trolleyfpeqconstfinal}
\end{equation}

\begin{equation}
    \left| \dot{d_T}(t) \right| \leq \overline{\dot{d_T}}, \quad \left| \ddot{d_T}(t) \right| \leq \overline{\ddot{d_T}(t)}
    \label{eq:trolleydTineqconst}
\end{equation}

\begin{equation}
    \left| \alpha_h(t) \right| \leq \overline{\alpha_h}, \quad \left| \alpha_l(t) \right| \leq \overline{\alpha_l}
    \label{eq:trolleyswingineqconst}
\end{equation}

In the aforementioned equations, $d_{Ti}$ and $d_{Tf}$ are the initial and final positions of the trolley along the jib, while $\overline{\dot{d_T}}$, $\overline{\ddot{d_T}}$, $\overline{\alpha_h}$ and $\overline{\alpha_l}$ are the maximum allowed values of the state DOFs. Also, expressions of the state variables are given in terms of the flat output $d_l$, via Eqs. (\ref{eq:trolleylfp} - \ref{eq:trolleydTaccfp}).

\subsection{Slew trajectory constraints}
\label{subsec:slewmotop}

Considering $t_S$ to be the total time for the slew operation, velocities and accelerations of the jib, and the hook-payload swings at the initial and final frames of the motion are to be constrained at zero value. The slew jerk must attain zero value at the start and end of the operation for smooth trajectory generation. As the jib rotates, its velocity and acceleration together with those of the resulting swings should be within their permissible limits.

Based on the expressions of the slew angle, the hook swing angles and the payload swing angles provided by Eqs. (\ref{eq:slewthetaSfp} - \ref{eq:slewblfp}), the corresponding constraints of the slew flat outputs can be obtained as:

\begin{equation}
    \begin{aligned}
        & x_h(0) = x_l(0) = D_TC_{\theta_{Si}}, \; \dot{x_h}(0) = \dot{x_l}(0) = 0, \\
        & \ddot{x_h}(0) = \ddot{x_l}(0) = 0, \\
        & x_l^{(3)}(0) = x_l^{(4)}(0) = x_l^{(5)}(0) = x_l^{(6)}(0) = x_l^{(7)}(0) = 0
    \end{aligned}
    \label{eq:slewfpxeqconstinitial}
\end{equation}

\begin{equation}
    \begin{aligned}
        & y_h(0) = y_l(0) = D_TS_{\theta_{Si}}, \; \dot{y_h}(0) = \dot{y_l}(0) = 0, \\
        & \ddot{y_h}(0) = \ddot{y_l}(0) = 0
    \end{aligned}
    \label{eq:slewfpyeqconstinitial}
\end{equation}

\begin{equation}
    \begin{aligned}
        & x_h(t_S) = x_l(t_S) = D_TC_{\theta_{Sf}}, \; \dot{x_h}(t_S) = \dot{x_l}(t_S) = 0, \\
        & \ddot{x_h}(t_S) = \ddot{x_l}(t_S) = 0, \\
        & x_l^{(3)}(t_S) = x_l^{(4)}(t_S) = x_l^{(5)}(t_S) = x_l^{(6)}(t_S) = x_l^{(7)}(t_S) = 0
    \end{aligned}
    \label{eq:slewfpxeqconstfinal}
\end{equation}

\begin{equation}
    \begin{aligned}
        & y_h(t_S) = y_l(t_S) = D_TS_{\theta_{Sf}}, \; \dot{y_h}(t_S) = \dot{y_l}(t_S) = 0, \\
        & \ddot{y_h}(t_S) = \ddot{y_l}(t_S) = 0
    \end{aligned}
    \label{eq:slewfpyeqconstfinal}
\end{equation}

\begin{equation}
    \left| \dot{\theta_S}(t) \right| \leq \overline{\dot{\theta_S}}, \quad \left| \ddot{\theta_S}(t) \right| \leq \overline{\ddot{\theta_S}(t)}
    \label{eq:slewthetaSineqconst}
\end{equation}

\begin{equation}
    \begin{aligned}
        & \left| \alpha_h(t) \right| \leq \overline{\alpha_h}, \quad \left| \beta_h(t) \right| \leq \overline{\beta_h}, \\
        & \left| \alpha_l(t) \right| \leq \overline{\alpha_l}, \quad \left| \beta_l(t) \right| \leq \overline{\beta_l}
    \end{aligned}
    \label{eq:slewswingineqconst}
\end{equation}

In these equations, $\theta_{Si}$ and $\theta_{Sf}$ are the initial and final angular positions of the crane jib, while $\overline{\dot{\theta_S}}$, $\overline{\ddot{\theta_S}}$, $\overline{\alpha_h}$, $\overline{\beta_h}$, $\overline{\alpha_l}$ and $\overline{\beta_l}$ are the maximum attainable values of the state DOFs. All the state variables are expressed in terms of the flat outputs $x_h$, $y_h$, $x_l$ and $y_l$, according to Eqs. (\ref{eq:slewthetaSfp} - \ref{eq:slewthetaSaccfp}).

\subsection{Flat output parameterization using B\'{e}zier curves}
\label{subsec:trajectoryparameterization}

The boundary conditions imposed on the trolley and slew operations can be utilized to parameterize the corresponding flat outputs with suitable polynomial curves. Noticing that there are a total of 16 boundary constraints involving each of $d_l$ and $x_l$, and 6 boundary conditions for each of $x_h$, $y_h$ and $y_l$, the individual flat outputs can be parameterized by $k$\textsuperscript{th} degree ($k=15 \; \textrm{for $d_l$ and $x_l$}, \; k=5 \; \textrm{for $x_h$, $y_h$ and $y_l$}$) B\'{e}zier curves \cite{LuigiBiagiotti2008}, which provide robust computational approach to compute the coefficients of the corresponding polynomial representations. Mathematically, the flat output position trajectories are expressed in the polynomial form as:

\begin{equation}
    \begin{aligned}
        p(t) & = \displaystyle\sum_{i=0}^{15} \tensor[^p]{c}{_i}\left( \frac{t}{t_{op}} \right)^{15}; \quad \forall p \in \{d_l, x_l\}, \; 0 \leq t \leq t_{op} \\
        & = \displaystyle\sum_{i=0}^{5} \tensor[^p]{c}{_i}\left( \frac{t}{t_{op}} \right)^{5}; \quad \forall p \in \{x_h, y_h, y_l\}, \; 0 \leq t \leq t_{op}
    \end{aligned}
    \label{eq:trajectorypoly}
\end{equation}

The scalar coefficients $\tensor[^p]{c}{_i}$ are given by:

\begin{equation}
    \begin{aligned}
        & \tensor[^{d_l}]{c}{_0} = d_{Ti}, \;\tensor[^{d_l}]{c}{_1} = 0, \; \tensor[^{d_l}]{c}{_2} = 0, \; \tensor[^{d_l}]{c}{_3} = 0, \\
        & \tensor[^{d_l}]{c}{_4} = 0, \; \tensor[^{d_l}]{c}{_5} = 0, \; \tensor[^{d_l}]{c}{_6} = 0, \; \tensor[^{d_l}]{c}{_7} = 0, \\
        & \tensor[^{d_l}]{c}{_8} = 6435\Delta d_T, \; \tensor[^{d_l}]{c}{_9} = -40040\Delta d_T, \\
        & \tensor[^{d_l}]{c}{_{10}} = 108108\Delta d_T, \; \tensor[^{d_l}]{c}{_{11}} = -163800\Delta d_T, \\
        & \tensor[^{d_l}]{c}{_{12}} = 150150\Delta d_T, \; \tensor[^{d_l}]{c}{_{13}} = -83160\Delta d_T, \\
        & \tensor[^{d_l}]{c}{_{14}} = 25740\Delta d_T, \; \tensor[^{d_l}]{c}{_{15}} = -3432\Delta d_T
    \end{aligned}
    \label{eq:trolleytrajectorycoeff}
\end{equation}

\begin{equation}
    \begin{aligned}
        & \tensor[^{x_l}]{c}{_0} = D_TC_{\theta_{Si}}, \;\tensor[^{x_l}]{c}{_1} = 0, \; \tensor[^{x_l}]{c}{_2} = 0, \; \tensor[^{d_l}]{c}{_3} = 0, \\
        & \tensor[^{x_l}]{c}{_4} = 0, \; \tensor[^{x_l}]{c}{_5} = 0, \; \tensor[^{x_l}]{c}{_6} = 0, \; \tensor[^{x_l}]{c}{_7} = 0, \\
        & \tensor[^{x_l}]{c}{_8} = 6435 D_T\Delta C_{\theta_S}, \; \tensor[^{x_l}]{c}{_9} = -40040 D_T\Delta C_{\theta_S}, \\
        & \tensor[^{x_l}]{c}{_{10}} = 108108 D_T\Delta C_{\theta_S}, \; \tensor[^{x_l}]{c}{_{11}} = -163800D_T\Delta C_{\theta_S}, \\
        & \tensor[^{x_l}]{c}{_{12}} = 150150 D_T\Delta C_{\theta_S}, \; \tensor[^{x_l}]{c}{_{13}} = -83160 D_T\Delta C_{\theta_S}, \\
        & \tensor[^{x_l}]{c}{_{14}} = 25740 D_T\Delta C_{\theta_S}, \; \tensor[^{x_l}]{c}{_{15}} = -3432 D_T\Delta C_{\theta_S}
    \end{aligned}
    \label{eq:slewxltrajectorycoeff}
\end{equation}

\begin{equation}
    \begin{aligned}
        & \tensor[^{x_h}]{c}{_0} = D_TC_{\theta_{Si}}, \; \tensor[^{x_h}]{c}{_1} = 0, \\
        & \tensor[^{x_h}]{c}{_2} = 0, \; \tensor[^{x_h}]{c}{_3} = 10 D_T\Delta C_{\theta_S}, \\
        & \tensor[^{x_h}]{c}{_4} = -15 D_T\Delta C_{\theta_S}, \; \tensor[^{x_h}]{c}{_5} = 6 D_T\Delta C_{\theta_S}
    \end{aligned}
    \label{eq:slewxhtrajectorycoeff}
\end{equation}

\begin{equation}
    \begin{aligned}
        & \tensor[^{y_h}]{c}{_0} = \tensor[^{y_l}]{c}{_0} = D_TS_{\theta_{Si}}, \; \tensor[^{y_h}]{c}{_1} = \tensor[^{y_l}]{c}{_1} = 0, \\
        & \tensor[^{y_h}]{c}{_2} = \tensor[^{y_l}]{c}{_2} = 0, \; \tensor[^{y_h}]{c}{_3} = \tensor[^{y_l}]{c}{_3} = 10 D_T\Delta S_{\theta_S}, \\
        & \tensor[^{y_h}]{c}{_4} = \tensor[^{y_l}]{c}{_4} = -15 D_T\Delta S_{\theta_S}, \; \tensor[^{y_h}]{c}{_5} = \tensor[^{y_l}]{c}{_5} = 6 D_T\Delta S_{\theta_S}
    \end{aligned}
    \label{eq:slewytrajectorycoeff}
\end{equation}

Here, $\Delta d_T = d_{Tf} - d_{Ti}$, $\Delta C_{\theta_S} = C_{\theta_{Sf}} - C_{\theta_{Si}}$, and $\Delta S_{\theta_S} = S_{\theta_{Sf}} - S_{\theta_{Si}}$. Substituting the polynomial coefficients in Eq. (\ref{eq:trajectorypoly}), the trolley and slew flat output trajectories are completely parameterized in terms of $t_T$ and $t_S$, respectively.

Following the aforementioned parameterization process, the MOTOPs for the trolley and slew operations can be formulated as parameterized multi-objective optimization problems with inequality constraints:

\begin{equation}
    \begin{aligned}
        \min _{t_T>0}\quad & \left[ f_1(t_T) \; f_2(t_T) \right]^\textrm{T} \\
        \textrm{s.t.} \quad & \textrm{Eqs. (\ref{eq:trolleydTineqconst} - \ref{eq:trolleyswingineqconst})}
    \end{aligned}
    \label{eq:trolleyfpMOTOP}
\end{equation}

\begin{equation}
    \begin{aligned}
        \min _{t_S>0}\quad & \left[ f_1(t_S) \; f_2(t_S) \right]^\textrm{T} \\
        \textrm{s.t.} \quad & \textrm{Eqs. (\ref{eq:slewthetaSineqconst} - \ref{eq:slewswingineqconst})}
    \end{aligned}
    \label{eq:slewfpMOTOP}
\end{equation}

Here, total operating times $f_1(t_T)$ and $f_1(t_S)$, and total operating energies (normalized actuator efforts) $f_2(t_T)$ and $f_2(t_S)$ are described according to:

\begin{equation}
    \begin{aligned}
        & f_1(t_T) = t_T, \\
        & f_2(t_T) = \displaystyle\int_0^{t_T} \left| \frac{\ddot{d_T}(t)}{\overline{\ddot{d_T}}} \right|^2 \, dt\\
    \end{aligned}
    \label{eq:trolleyMOTOPobj}
\end{equation}

\begin{equation}
    \begin{aligned}
        & f_1(t_S) = t_S, \\
        & f_2(t_S) = \displaystyle\int_0^{t_S} \left| \frac{\ddot{\theta_S}(t)}{\overline{\ddot{\theta_S}}} \right|^2 \, dt\\
    \end{aligned}
    \label{eq:slewMOTOPobj}
\end{equation}

It is to be noted that in the above set of equations $\ddot{d_T}(t)$ and $\ddot{\theta_S}(t)$ are provided by Eq. (\ref{eq:trolleydTaccfp}) and Eq. (\ref{eq:slewthetaSaccfp}), respectively.

By solving the operating time $t_T$ and $t_S$ from Eqs. (\ref{eq:trolleyfpMOTOP} - \ref{eq:slewfpMOTOP}), the flat output trajectories $d_l(t)$, $x_h(t)$, $y_h(t)$, $x_l(t)$, and $y_l(t)$ can be computed according to Eq. (\ref{eq:trajectorypoly}). Subsequently, the trolley and slew reference trajectories along with the corresponding swing trajectories can be derived using Eqs. (\ref{eq:trolleylfp}-\ref{eq:trolleydTaccfp}) and (\ref{eq:slewthetaSfp} - \ref{eq:slewthetaSaccfp}).

\section{Collective Oppositional GDE3 for Multi-objective Trajectory Optimization}
\label{sec:COGDE3formotop}

Kukkonen and Lampinen developed GDE3 \cite{Kukkonen2005} to address global optimization problems with an arbitrary number of objectives and constraints. It is designed with the principles of simplicity, efficiency, and flexibility, for minimizing non-differentiable and nonlinear continuous functions. Since the MOTOPs constructed in Section \ref{sec:motop} have conflicting continuous objectives, with nonlinearity in the energy objectives, GDE3 is well-suited for the current scenario. In order to address the shortcomings of GDE3 in terms of initial population sensitivity, the following section constructs a new population seeding technique.

\subsection{Population initialization using computational opposition}
\label{subsec:CO}

The concept of computational opposition was first introduced by Tizhoosh \cite{Tizhoosh2005} as an efficient technique to improve the performance of various machine intelligence algorithms. Opposition and its different variants exploit the fact that for a given problem, simultaneous computation of a solution and its opposite provides a better chance of finding a candidate solution nearer to the global optimum. Some of these variants found in literature are opposition \cite{Tizhoosh2005}, quasi-opposition \cite{Rahnamayan2007}, quasi-reflection \cite{Ergezer2009}, extended opposition \cite{Seif2015}, reflected extended opposition \cite{Seif2015} and others. The definitions of the various opposites of a solution in one-dimensional (1D) space, as demonstrated in Fig. \ref{fig:oppositepoints}, are given below.

\begin{figure}[!t]
    \begin{center}   
        \includegraphics[width=\columnwidth]{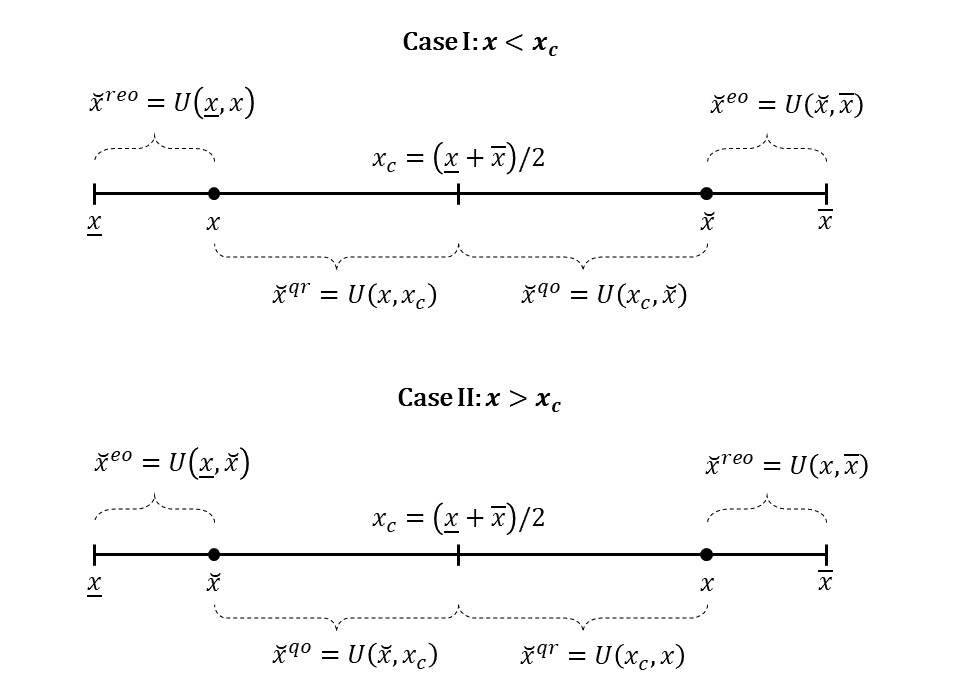}
        \caption{Positions and ranges of various opposites (opposite solution $\breve{x}$, quasi-opposite solution $\breve{x}^{qo}$, quasi-reflected solution $\breve{x}^{qr}$, extended opposite solution $\breve{x}^{eo}$, reflected extended opposite solution $\breve{x}^{reo}$) of a random solution $x$ within the interval $[\underline{x},\overline{x}]$. $U(l,u)$ indicates a randomly sampled number following a uniform probability distribution within $[l,u]$. Two possible cases are shown with the position of $x$ lying before and after the center of its domain.}
        \label{fig:oppositepoints}
    \end{center}
\end{figure}

\begin{definition}
    \textbf{(Opposite solution \cite{Tizhoosh2005})} For a solution $x \in [\underline{x},\overline{x}]$ in 1D space, where $\underline{x},\overline{x} \in \mathbb{R}$, the opposite solution $\breve{x}$ is the reflection of $x$ about the center of the solution space. It is denoted by:
    
    \begin{equation}
        \breve{x} = \underline{x} + \overline{x} - x
        \label{eq:opppoint}
    \end{equation}
\end{definition}

\begin{definition}
    \textbf{(Quasi-opposite solution \cite{Rahnamayan2007})} Let $U(l,u)$ indicate a randomly sampled number following a uniform probability distribution within $[l,u]$. For a solution $x \in [\underline{x},\overline{x}]$ in 1D space, where $\underline{x},\overline{x} \in \mathbb{R}$, the quasi-opposite solution $\breve{x}^{qo}$ is the reflection of $x$ to a random point between the center of the solution space and $\breve{x}$. It is denoted by:
    
    \begin{equation}
        \breve{x}^{qo} = U\left(\frac{\underline{x}+\overline{x}}{2},\breve{x}\right)
        \label{eq:quasiopppoint}
    \end{equation}
\end{definition}

\begin{definition}
    \textbf{(Quasi-reflected solution \cite{Ergezer2009})} For a solution $x \in [\underline{x},\overline{x}]$ in 1D space, where $\underline{x},\overline{x} \in \mathbb{R}$, the quasi-reflected solution $\breve{x}^{qo}$ is the reflection of $x$ to a random point between the center of the solution space and $x$. It is denoted by:
    
    \begin{equation}
        \breve{x}^{qr} = U\left(x,\frac{\underline{x}+\overline{x}}{2}\right)
        \label{eq:quasirefpoint}
    \end{equation}
\end{definition}

\begin{definition}
    \textbf{(Extended opposite solution \cite{Seif2015})} For a solution $x \in [\underline{x},\overline{x}]$ in 1D space, where $\underline{x},\overline{x} \in \mathbb{R}$, the extended opposite solution $\breve{x}^{eo}$ is the reflection of $x$ to a random point between $\breve{x}$ and the bound of the solution space nearest to $\breve{x}$. It is denoted by:
    
    \begin{equation}
        \begin{aligned}
            \breve{x}^{eo} & = U\left(\breve{x},\overline{x}\right); \quad x < \frac{\underline{x}+\overline{x}}{2} \\
            & = U\left(\underline{x},\breve{x}\right); \quad x > \frac{\underline{x}+\overline{x}}{2}
        \end{aligned}
        \label{eq:extopppoint}
    \end{equation}
\end{definition}

\begin{definition}
    \textbf{(Reflected extended opposite solution \cite{Seif2015})} For a solution $x \in [\underline{x},\overline{x}]$ in 1D space, where $\underline{x},\overline{x} \in \mathbb{R}$, the reflected extended opposite solution $\breve{x}^{reo}$ is the reflection of $x$ to a random point between $x$ and the bound of the solution space nearest to $x$. It is denoted by:
    
    \begin{equation}
        \begin{aligned}
            \breve{x}^{reo} & = U\left(\underline{x},x\right); \quad x < \frac{\underline{x}+\overline{x}}{2} \\
            & = U\left(x,\overline{x}\right); \quad x > \frac{\underline{x}+\overline{x}}{2}
        \end{aligned}
        \label{eq:refextopppoint}
    \end{equation}
\end{definition}

According to \cite{Rahnamayan2007} and \cite{Seif2015}, in 1D solution space, the expected values of the probabilities of the different opposites of a random solution $x$ being closer to an unknown optimum $x^*$, compared to $x$, are derived as follows:

\begin{equation}
    \begin{aligned}
        & E\left(Pr\left(|\breve{x}-x^*| < |x-x^*|\right)\right) = \frac{1}{2} \\
        & E\left(Pr\left(|\breve{x}^{qo}-x^*| < |x-x^*|\right)\right) = \frac{9}{16} \\
        & E\left(Pr\left(|\breve{x}^{qr}-x^*| < |x-x^*|\right)\right) = \frac{11}{16} \\
        & E\left(Pr\left(|\breve{x}^{eo}-x^*| < |x-x^*|\right)\right) = \frac{7}{16} \\
        & E\left(Pr\left(|\breve{x}^{reo}-x^*| < |x-x^*|\right)\right) = \frac{3}{16} \\
    \end{aligned}
    \label{eq:probopppoints}
\end{equation}

Here, $E()$ is the expected value function and $Pr()$ is the probability function.

By observing Fig. \ref{fig:oppositepoints} and analyzing Eq. (\ref{eq:probopppoints}), it can be inferred that the collective set of opposites of a random solution provides a trade-off between exploration and exploitation for a multi-objective optimization problem. They span the whole solution space, while overall possessing higher chances of finding a Pareto optimal point in the solution landscape. Hence, if the initial population of an MOEA can be seeded with such opposites of randomly generated solutions, then \textit{all} the potentially \textit{good} portions of the solution landscape can be traversed through successive generations.

In light of the foregoing analysis, a novel collective oppositional population initialization approach is integrated into the GDE3 procedure \cite{Kukkonen2005}, as presented in Algorithm \ref{algo:coinit}. A set of random solutions are generated within the solution domain, with cardinality equal to a fifth of the required size of the initial population. Then the various opposites of each of those solutions are computed using Eqs. (\ref{eq:opppoint} - \ref{eq:refextopppoint}). Together, all five types of opposite solutions are named \textit{collective opposite solutions}. Subsequently, the initial population is prepared by combining all the collective opposite solutions. This initialization technique is aimed towards creating a population that can cover the whole search space with a higher number of members closer to the Pareto optimal solutions, compared to those of a randomly sampled population. 

\begin{algorithm}[tbp]
    \begin{algorithmic}[1]
        \STATE \textbf{input} Population size $\mathcal{N}$, Lower bound of solution space $\underline{x}$, Upper bound of solution space $\overline{x}$
        \STATE \textbf{output} Population $\mathbf{P}_0$
        \STATE \textbf{initialize} Random population $\mathbf{P}$ with $|\mathbf{P}| = \mathcal{N}/5$, Empty populations $\mathbf{P}_o$, $\mathbf{P}_{qo}$, $\mathbf{P}_{qr}$, $\mathbf{P}_{eo}$ and $\mathbf{P}_{reo}$
        \FOR{$i$ = $1$ to $\mathcal{N}/5$}
            \STATE Compute $\breve{x}_i$ from $x_i$ in $\mathbf{P}$ according to Eq. (\ref{eq:opppoint}) and set $\mathbf{P}_o = \mathbf{P}_o \cup \{\breve{x}_i\}$;
            \STATE Compute $\breve{x}^{qo}_i$ from $\breve{x}_i$ in $\mathbf{P}_o$ according to Eq. (\ref{eq:quasiopppoint}) and set $\mathbf{P}_{qo} = \mathbf{P}_{qo} \cup \{\breve{x}^{qo}_i\}$;
            \STATE Compute $\breve{x}^{qr}_i$ from $x_i$ in $\mathbf{P}$ according to Eq. (\ref{eq:quasirefpoint}) and set $\mathbf{P}_{qr} = \mathbf{P}_{qr} \cup \{\breve{x}^{qr}_i\}$;
            \STATE Compute $\breve{x}^{eo}_i$ from $\breve{x}_i$ in $\mathbf{P}_o$ according to Eq. (\ref{eq:extopppoint}) and set $\mathbf{P}_{eo} = \mathbf{P}_{eo} \cup \{\breve{x}^{eo}_i\}$;
            \STATE Compute $\breve{x}^{reo}_i$ from $x_i$ in $\mathbf{P}$ according to Eq. (\ref{eq:refextopppoint}) and set $\mathbf{P}_{reo} = \mathbf{P}_{reo} \cup \{\breve{x}^{reo}_i\}$;
        \ENDFOR
        \STATE Set $\mathbf{P}_0 = \mathbf{P}_o \cup \mathbf{P}_{qo} \cup \mathbf{P}_{qr} \cup \mathbf{P}_{eo} \cup \mathbf{P}_{reo}$;
        \RETURN $\mathbf{P}_0$
    \end{algorithmic}
\caption{Collective Oppositional Population Initialization}
\label{algo:coinit}
\end{algorithm}

\subsection{CO-GDE3 procedure}
\label{subsec:COGDE3}

\begin{figure}[tbp]
    \begin{center}
        \includegraphics[width=\columnwidth]{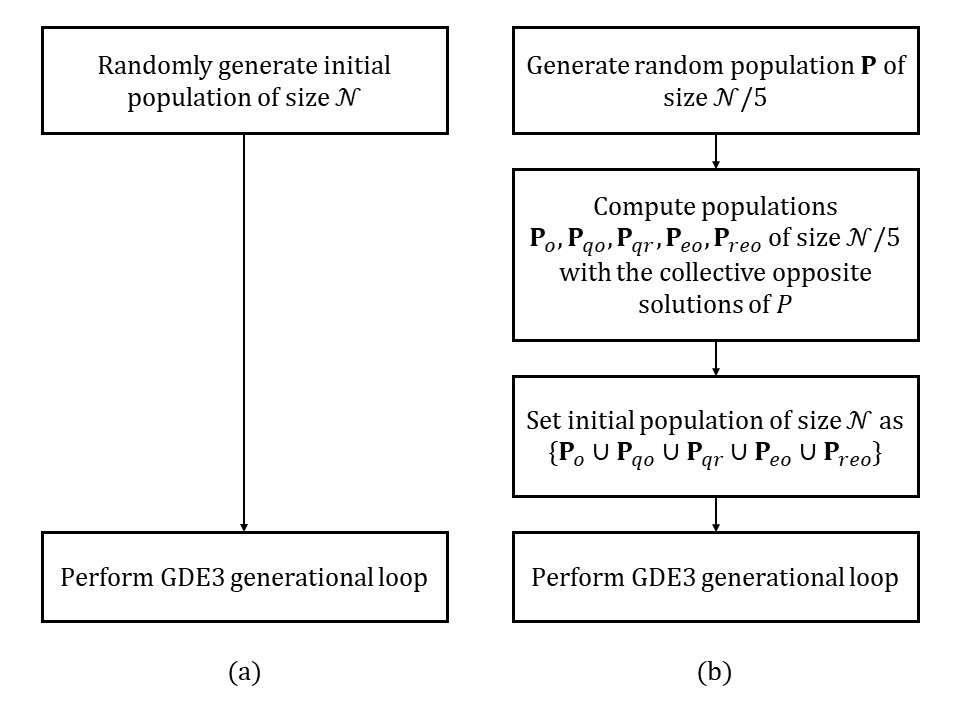}
        \caption{Algorithmic flowchart of (a) GDE3 and (b) CO-GDE3.}
        \label{fig:COGDE3}
    \end{center}
\end{figure}

Using the proposed initialization technique, the effectiveness and efficiency of the conventional GDE3 \cite{Kukkonen2005} can be improved, by enhancing the quality of the initial population. The algorithmic procedure of the CO-GDE3 is illustrated in Fig. \ref{fig:COGDE3}, in comparison to that of the standard GDE3. The design principles of the CO-GDE3 algorithm are along the line of those of the standard GDE3. In contrast to \cite{HernandezDiaz2008,Liu2015,Friedrich2015,Bhowmik2015,Ewees2021}, the initialization strategy is kept simple and computationally efficient, free of costly objective function evaluations of the solutions or additional parameters. The computational complexity of the proposed initialization strategy is solely dependent on that of the random population generation of Line 3 in Algorithm \ref{algo:coinit}. The worst-case complexity for such an operation using a uniform random number generator is $O(N)$ where $N$ is the size of the population. In the proposed initialization algorithm, to generate an initial population $P_0$ with $\mathcal{N}$ solutions, population $P$ of size $\mathcal{N}/5$ is computed with complexity $O(\mathcal{N})$. Each of the individual population $P_o, P_{qo}, P_{qr}, P_{eo}, P_{reo}$ is obtained with exactly $\mathcal{N}/5$ operations. Hence the overall worst-case complexity of the collective oppositional initialization algorithm is $O(\mathcal{N})$. For a standard GDE3, where $\mathcal{N}$ solutions are evolved through $G_{max}$ generations to optimize $M$ objective functions, the total worst-case complexity is given by $O(G_{max}\mathcal{N}\log^{M-1}\mathcal{N})$. So, the CO-GDE3 also has the same worst-case computational complexity, and Algorithm \ref{algo:coinit} does not add any computational burden to the conventional GDE3. Using CO-GDE3, the MOTOPs of the current work can be effectively solved.

\subsection{Selection of optimal solution from CO-GDE3 output}
\label{subsec:AFMFselection}

For both the trolley and the slew MOTOPs discussed in this research, the weighting of the objective values (operating time and energy), denoting the preference of the crane controller, usually varies in a fuzzy manner based on the scenario of the problem. To address this, an average fuzzy membership function (AFMF) is employed as the multi-criteria decision-maker to pick a single solution as the operating condition from the generated $\mathcal{PF}_{known}$. The linear membership functions for the objective values $f_1$ and $f_2$ of a solution $x^*_i$ from $\mathcal{PF}_{known}$ is constructed as:

\begin{equation}
    \lambda_j(x^*_i) = \frac{f_{jmax}-f_j(x^*_i)}{f_{jmax}-f_{jmin}}; \quad \forall i \in \{ 1,2,3,...,\mathcal{N} \}, j = 1,2
    \label{eq:lambda}
\end{equation}

$\lambda_j(x^*_i)$ becomes 1 when $x^*_i$ corresponds to $f_{jmin}$, for $j=1,2$. For the bi-objective minimization problem of the tower crane trajectory planning task with conflicting objectives, both these cases are unattainable simultaneously. Therefore, the average of the linear fuzzy membership functions is utilized as the evaluation function to select a solution from the Pareto set.

\begin{equation}
    \Lambda(x^*_i) = \frac{1}{2} \displaystyle\sum_{j=0}^2\lambda_j(x^*_i); \quad \forall i \in \{ 1,2,3,...,\mathcal{N} \}
    \label{eq:lambdabar}
\end{equation}

The solution with the highest evaluation function value $\Lambda_{max}$ is considered to be the optimal solution to the MOTOP.

\begin{figure}[tp]
    \begin{center}
        \includegraphics[width=\columnwidth]{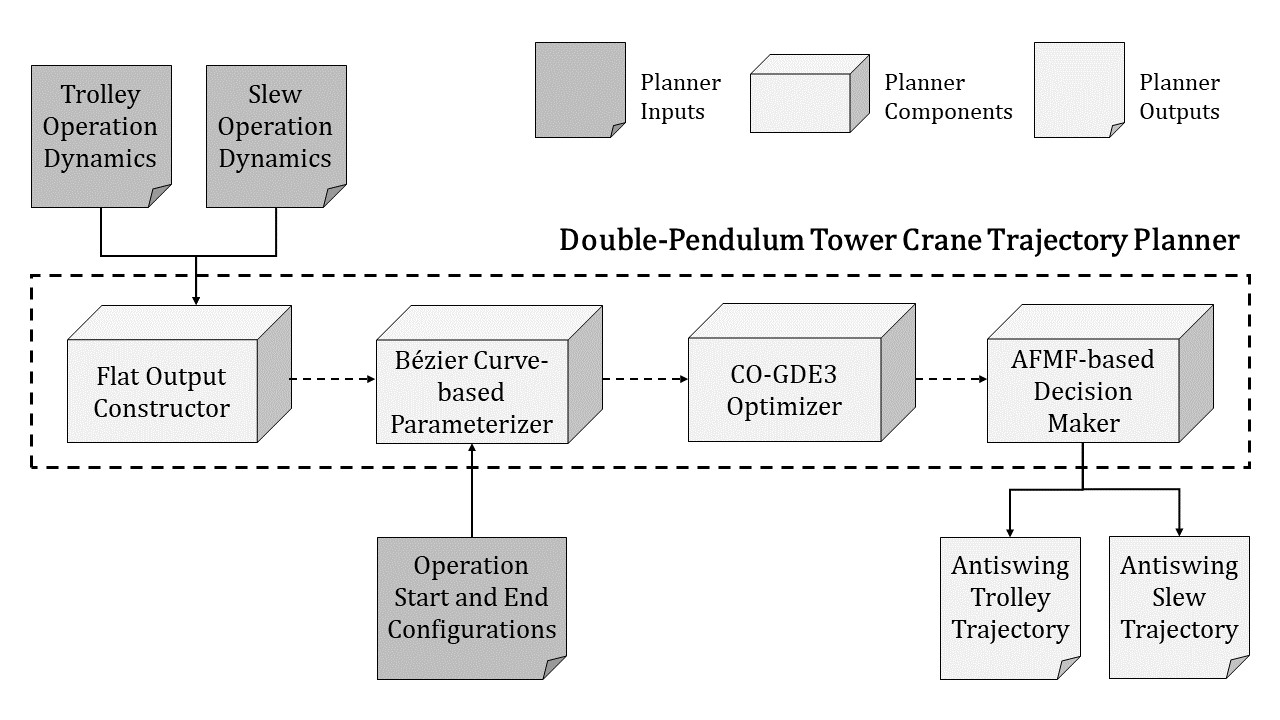}
        \caption{Anti-swing trajectory planning architecture for autonomous tower cranes with double-pendulum dynamics.}
        \label{fig:trajectoryplanner}
    \end{center}
\end{figure}

The complete architecture of the developed anti-swing trajectory planner for double-pendulum tower cranes is illustrated in Fig. \ref{fig:trajectoryplanner}. The primary components of the planner are the flat output constructor, the B\'{e}zier curve-based trajectory parameterizer, the CO-GDE3 optimizer and the AFMF-based decision-maker.

\section{Simulation Studies}
\label{sec:simulations}

To evaluate the performance of the proposed trajectory planning method, simulations are conducted in MATLAB R2021a environment, on a PC with a 2.20 GHz Intel(R) Core(TM) i7-8750H CPU and 8 GB installed physical memory (RAM). The MOEAs are employed via PlatEMO v3.2 \cite{Tian2017}. An NVIDIA GeForce GTX 1650 display adapter with 896 NVIDIA CUDA cores and 4 GB standard memory configuration is utilized as the GPU.

\subsection{Procedure}
\label{subsec:procedure}

\begin{figure}[tp]
    \begin{center}
        \includegraphics[width=\columnwidth]{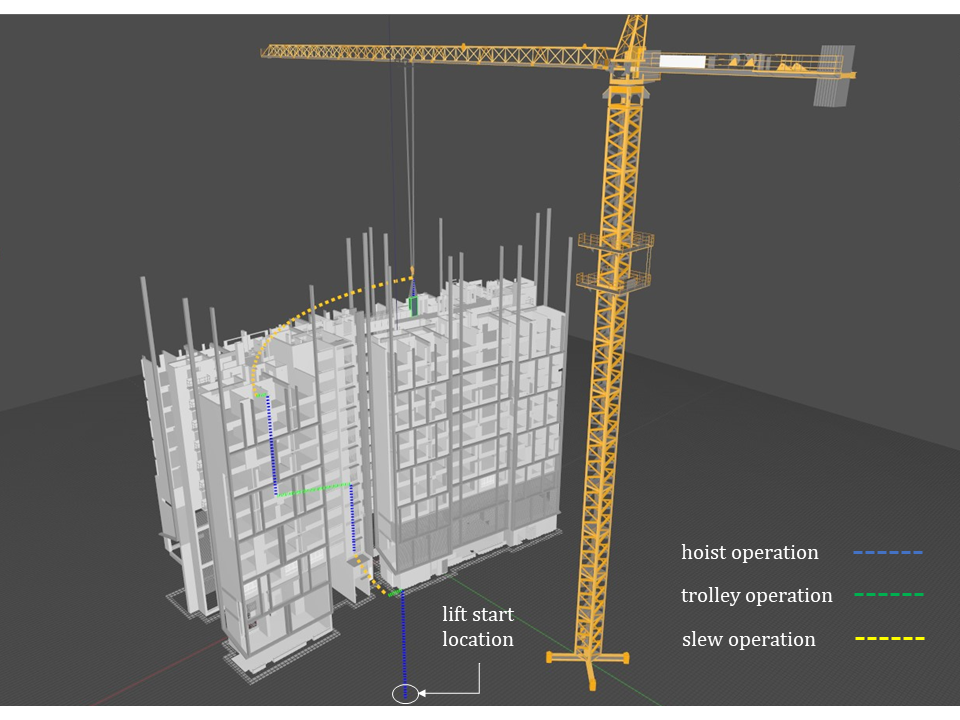}
        \caption{Collision-free lifting path from the path planner of the CALP system used in the trajectory planning simulation studies.}
        \label{fig:liftingpath}
    \end{center}
\end{figure}

Simulations are performed using a 1:10 scaled virtual model of Terex SK 415-20 hammerhead tower crane in the CALP system \cite{Cai2016,Dutta2020} developed at Nanyang Technological University. Limits on the tower crane motion given in TABLE \ref{tbl:craneparameters} are utilized to perform trajectory planning for the operations listed in TABLE \ref{tbl:operations}. The operational values (boundary positions) are obtained from portions of the collision-free lifting path (shown in Fig. \ref{fig:liftingpath}) provided by the path planner of the CALP system. The maximum swing angle values are selected based on the maximum allowed deflection of the payload from the planned lifting path, according to the threshold distance allowed by the multi-level Oriented Bounding Boxes (OBBs) of the payload \cite{Dutta2020}, which ensure avoidance of collision with obstacles. The values are also kept within the small-angle approximation ranges ($< 5^\circ$).

\begin{table}[!t]
    \centering
    \caption{Mechanical properties and constraints of the tower crane used in the simulation studies.}
    \label{tbl:craneparameters}
    \begin{tabular}{clc}
        \hline
        \bf Parameter & \bf Meaning & \bf Value \\
        \hline
        $m_h$ & mass of hook & 0.5kg \\
        $m_l$ & mass of payload & 1kg \\
        $D_l$ & length of rig-cable & 1.5m \\
        $\overline{\dot{d_T}}$ & max trolley velocity & 0.5m/s \\
        $\overline{\ddot{d_T}}$ & max trolley acceleration & 0.5m/s\textsuperscript2 \\
        $\overline{\dot{\theta_S}}$ & max slew velocity & 20$^\circ$/s \\
        $\overline{\ddot{\theta_S}}$ & max slew acceleration & 20$^\circ$/s\textsuperscript2 \\
        $\overline{\alpha_h}$ & max radial hook swing angle & 2.5$^\circ$ \\
        $\overline{\beta_h}$ & max tangential hook swing angle & 2.5$^\circ$ \\
        $\overline{\alpha_l}$ & max radial payload swing angle & 2.5$^\circ$ \\
        $\overline{\beta_l}$ & max tangential payload swing angle & 2.5$^\circ$ \\
        $g$ & gravitational acceleration & 9.8m/s\textsuperscript2 \\
        \hline
    \end{tabular}
\end{table}

\begin{table}[!t]
    \centering
    \caption{Tower crane operations for the MOTOPs in the simulation studies.}
    \label{tbl:operations}
    \begin{tabular}{p{0.2\columnwidth}clc}
        \hline
        \bf Operation (Problem) & \bf Parameter & \bf Meaning & \bf Value \\
        \hline
        \multirow{5}{0.2\columnwidth}{Trolley \\ (T-MOTOP)} & $d_{Ti}$ & initial trolley position & 1m \\
        & $d_{Tf}$ & final trolley position & 2m \\
        & $D_h$ & fixed length of hoist-cable & 3m \\
        & $\overline{t_T}$ & max trolley operating time & 8s \\
        \hline
        \multirow{6}{0.2\columnwidth}{Slew \\ (S-MOTOP)} & $\theta_{Si}$ & initial slew angle & 30$^\circ$ \\
        & $\theta_{Sf}$ & final slew angle & 60$^\circ$ \\
        & $D_h$ & fixed length of hoist-cable & 3m \\
        & $D_T$ & fixed position of trolley & 1m \\
        & $\overline{t_S}$ & max slew operating time & 8s \\
        \hline
    \end{tabular}
\end{table}

The trolley and slew operations provide the two trajectory optimization problems (T-MOTOP and S-MOTOP) for the simulations.  The conventional GDE3 and CO-GDE3 are applied to the two MOTOPs to analyze and compare the performances of the algorithms for the double-pendulum tower crane trajectory planning problem. Both the MOEAs are configured with the parameters provided in TABLE \ref{tbl:GDE3parameters}. The termination criterion for the MOEAs is set as the maximum number of function evaluations by the algorithms. When the specified value is reached, the MOEA terminates and provides the latest Pareto front as the $\mathcal{PF}_{known}$. Keeping the probabilistic nature of MOEAs in consideration, 25 simulation runs for each of the MOTOPs are performed with each of the conventional GDE3 and the CO-GDE3.

\begin{table}[!t]
    \centering
    \caption{Parameter configurations for GDE3 and CO-GDE3 in the simulation studies.}
    \label{tbl:GDE3parameters}
    \begin{tabular}{clc}
        \hline
        \bf Parameter & \bf Meaning & \bf Value \\
        \hline
        $CR$ & crossover rate & 0.9 \\
        $F$ & scaling factor & 0.5 \\
        $\mathcal{N}$ & population size & 100 \\
        $FE_{max}$ & maximum number of function evaluations & 1000 \\
        \hline
    \end{tabular}
\end{table}

To measure and compare the effectiveness and efficiency of GDE3 and CO-GDE3, when applied to the two MOTOPs, the metrics considered in the current study are Hyperarea ($HA$) \cite{Coello2007}, Spacing ($SP$) \cite{Coello2007}, and Minimum Number of Function Evaluations ($FE_{min}$). The scales of comparison are selected by considering the general objectives of multi-objective optimization - finding $\mathcal{PF}_{known}$ as close to $PF_{true}$ as possible, providing a Pareto front with a wide range of objective values, and converging to $\mathcal{PF}_{known}$ as fast as possible, respectively.

Following the detailed comparison of the proposed optimizer to that of the novel optimizer, the trajectories obtained through the proposed anti-swing trajectory planner are analyzed to verify its effectiveness in obtaining multi-objective optimal operation trajectories, addressing the state constraints and reducing the residual double-swing.

\subsection{Results and analysis}
\label{subsec:results}

\subsubsection{Statistical comparison of GDE3 and CO-GDE3}
\label{subsubsec:MOEAresults}

\begin{figure}[!t]
    \centering
    \subfloat[Trolley MOTOP\label{subfig:trolleyhyperareaplot}]{%
        \centering
        \includegraphics[width=\columnwidth]{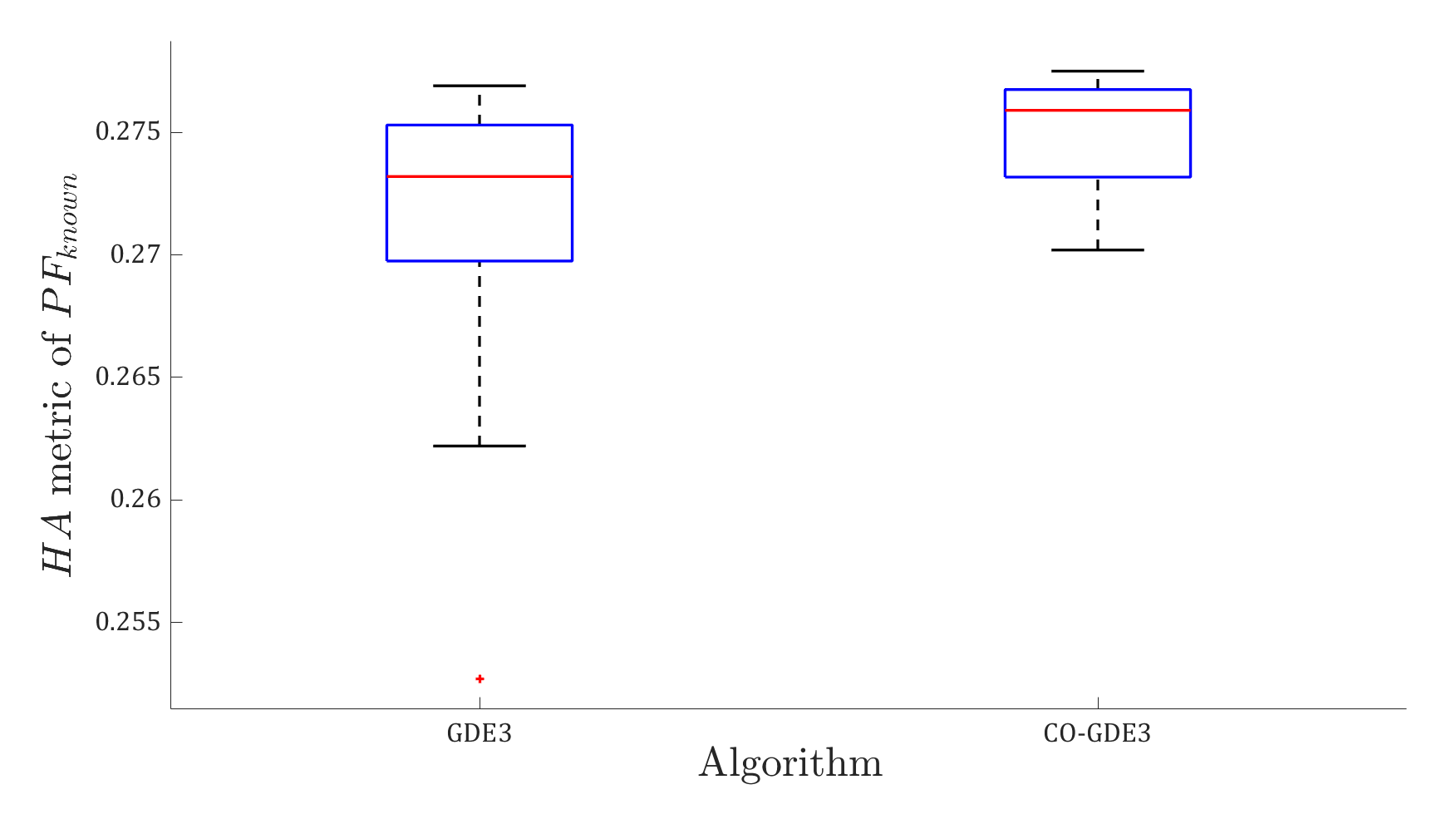}}
    \\
    \subfloat[Slew MOTOP\label{subfig:slewhyperareaplot}]{%
        \centering
        \includegraphics[width=\columnwidth]{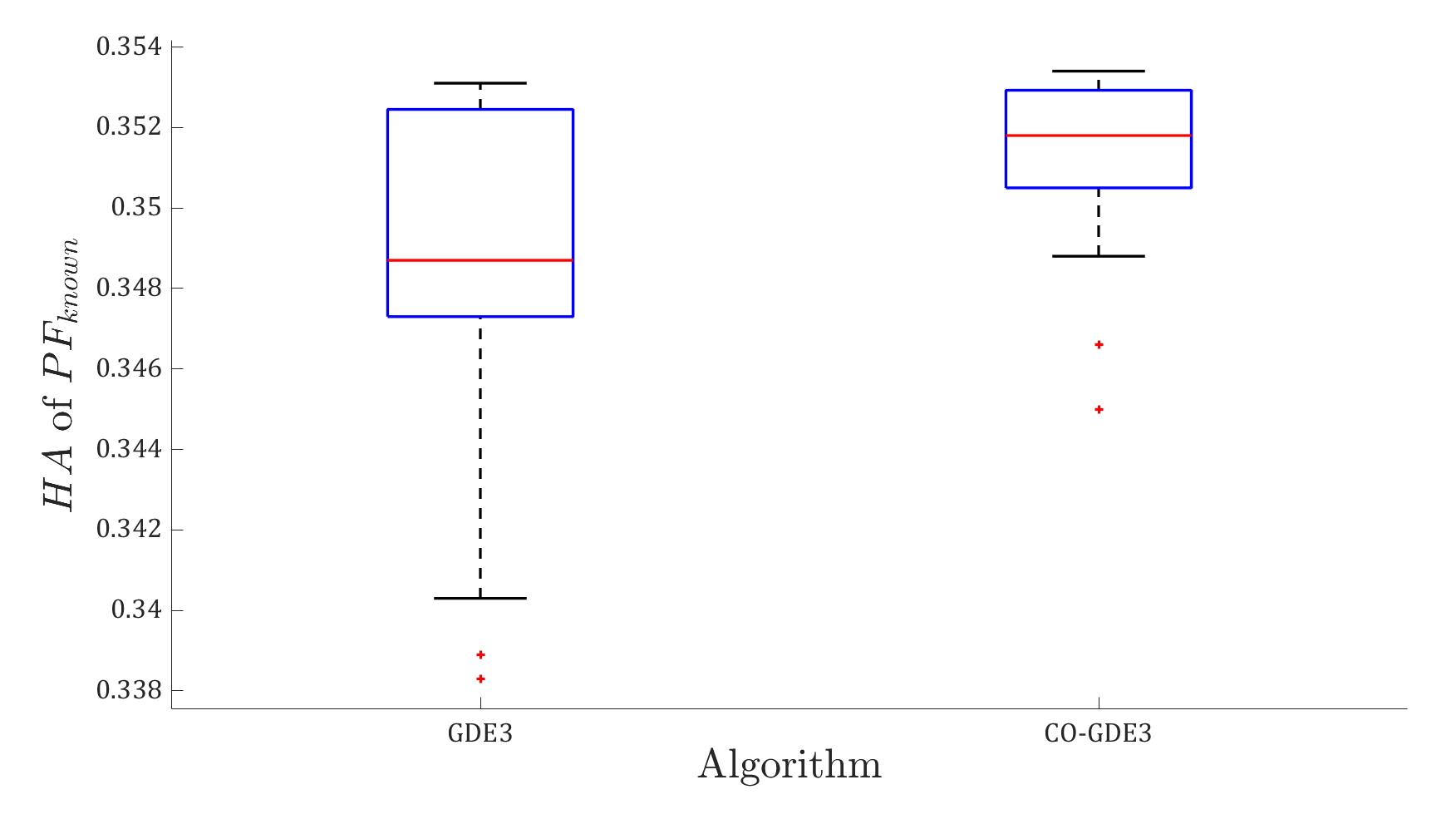}}
    \caption{Statistical measures of $HA$ metric of GDE3 and CO-GDE3 for (a) the T-MOTOP and (b) the S-MOTOP.}
    \label{fig:hyperareaplots}
\end{figure}

\begin{figure}[!t]
    \centering
    \subfloat[Trolley MOTOP\label{subfig:trolleyspacingplot}]{%
        \centering
        \includegraphics[width=\columnwidth]{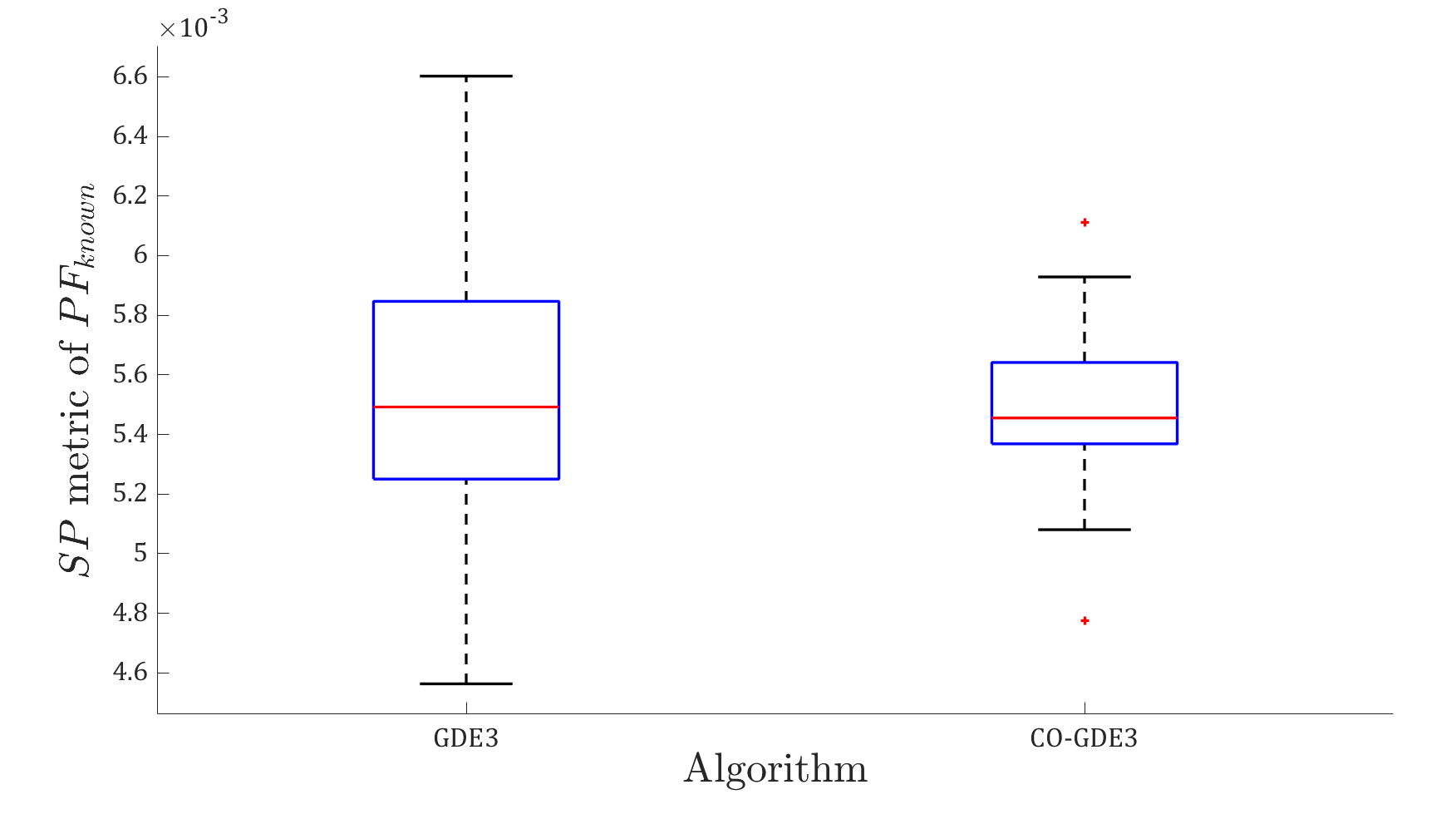}}
    \\
    \subfloat[Slew MOTOP\label{subfig:slewspacingplot}]{%
        \centering
        \includegraphics[width=\columnwidth]{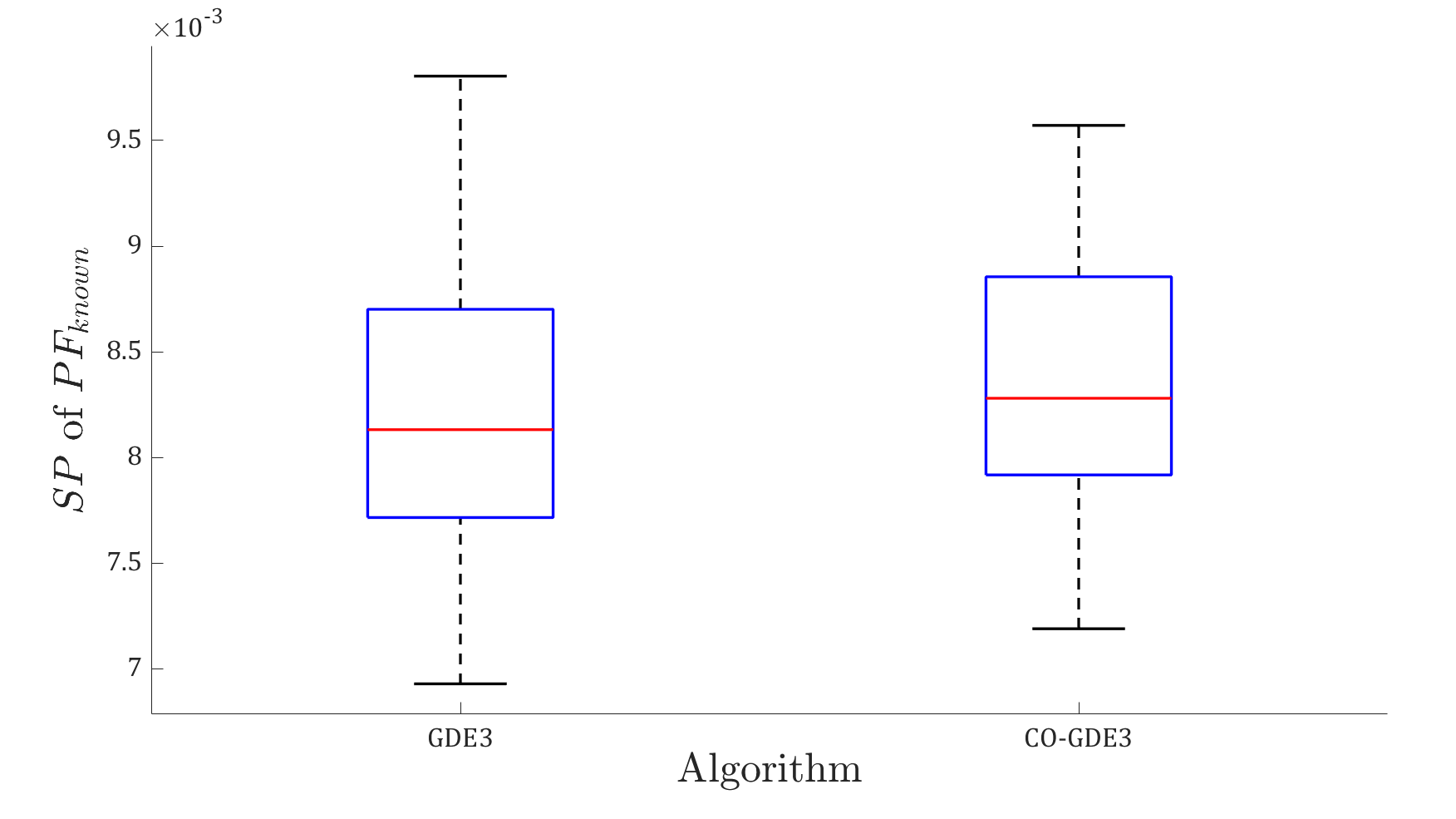}}
    \caption{Statistical measures of $SP$ metric of GDE3 and CO-GDE3 for (a) the T-MOTOP and (b) the S-MOTOP.}
    \label{fig:spacingplots}
\end{figure}

\begin{figure}[!t]
    \centering
    \subfloat[Trolley MOTOP\label{subfig:trolleyconvergenceplot}]{%
        \centering
        \includegraphics[width=\columnwidth]{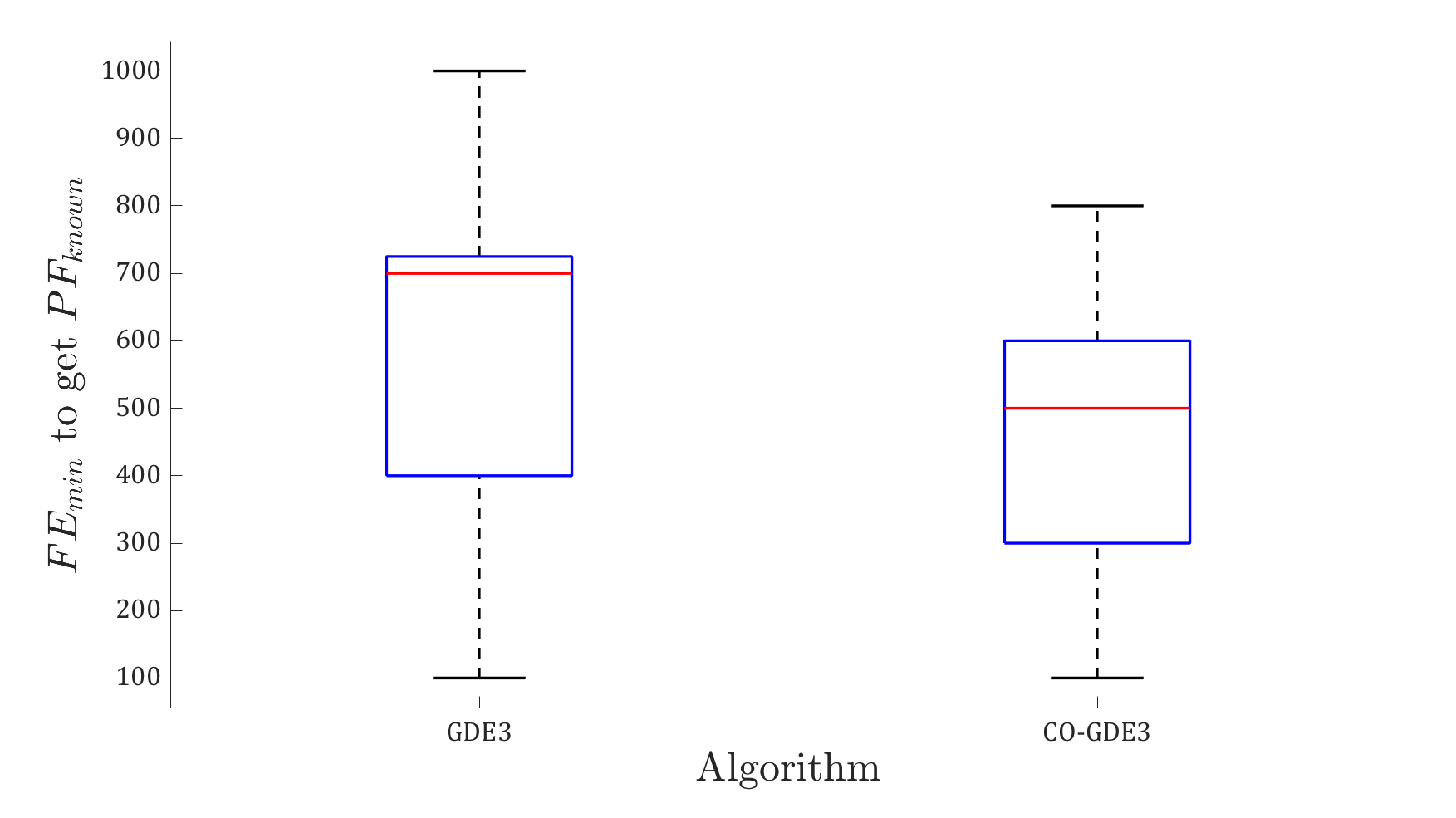}}
    \\
    \subfloat[Slew MOTOP\label{subfig:slewconvergenceplot}]{%
        \centering
        \includegraphics[width=\columnwidth]{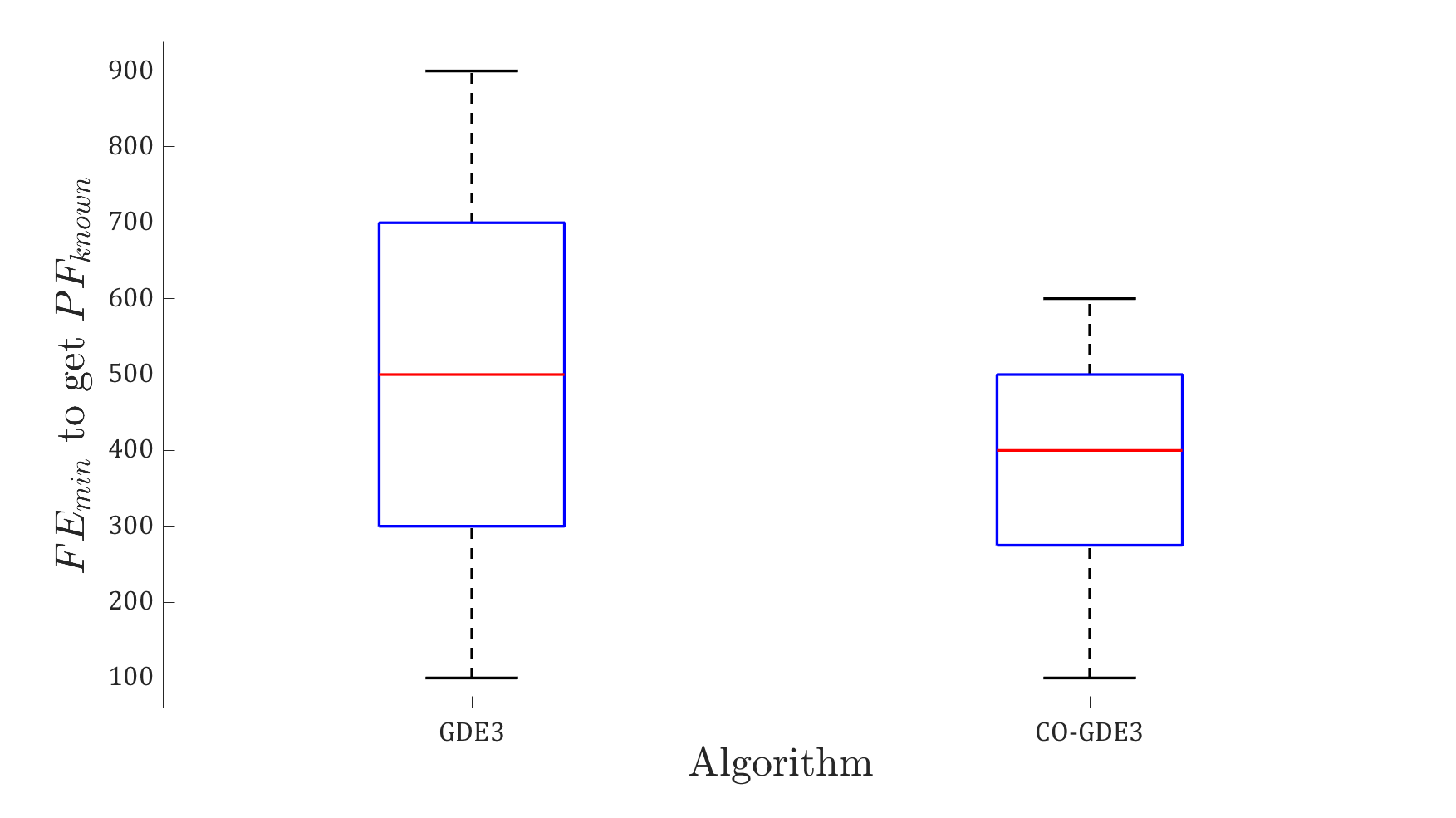}}
    \caption{Statistical measures of $FE_{min}$ metric of GDE3 and CO-GDE3 for (a) the T-MOTOP an (b) the S-MOTOP.}
    \label{fig:convergenceplots}
\end{figure}

For the T-MOTOP and the S-MOTOP, the statistical measures of the performance metrics of the two MOEAs in terms of $HA$, $SP$ and $FE_{min}$, for 25 runs on each MOTOP, are demonstrated in Fig. \ref{fig:hyperareaplots}, Fig. \ref{fig:spacingplots} and Fig. \ref{fig:convergenceplots}, respectively. The boxplots in these figures consist of five values: the maximum and minimum values, the lower and upper quartiles, and the median. These values are illustrated in ascending order - minimum value (Q0), lower quartile (Q1), median value (Q2), upper quartile (Q3), and maximum value (Q4). The corresponding mean and standard deviation values are compiled in TABLE \ref{tbl:trolleystats} for T-MOTOP and TABLE \ref{tbl:slewstats} for S-MOTOP.

\begin{table}[!t]
    \centering
    \caption{Statistical results of performance metrics of GDE3 and CO-GDE3 for the T-MOTOP of the simulation studies. (Better results of the metrics are highlighted in bold.)}
    \label{tbl:trolleystats}
    \begin{tabular}{ccccc}
        \hline 
        \multirow{2}{*}{\bf Metric} & \multicolumn{2}{c}{\textbf{Mean} ($\bm{\mu}$)} & \multicolumn{2}{c}{\textbf{Standard Deviation} ($\bm{\sigma}$)} \\
        \cline{2-5}
        & \bf GDE3 & \bf CO-GDE3 & \bf GDE3 & \bf CO-GDE3 \\
        \hline
        $HA$ & 2.71e-01 & \textbf{2.75e-01} & 5.50e-03 & \textbf{2.30e-03} \\
        $SP$ & 5.50e-03 & 5.50e-03 & 4.43e-04 & \textbf{2.82e-04} \\
        $FE_{min}$ & 592 & \textbf{456} & 239.65 & \textbf{198.07} \\
        \hline
    \end{tabular}
\end{table}

\begin{table}[!t]
    \centering
    \caption{Statistical results of performance metrics of GDE3 and CO-GDE3 for the S-MOTOP of the simulation studies. (Better results of the metrics are highlighted in bold.)}
    \label{tbl:slewstats}
    \begin{tabular}{ccccc}
        \hline
        \multirow{2}{*}{\bf Metric} & \multicolumn{2}{c}{\textbf{Mean} ($\bm{\mu}$)} & \multicolumn{2}{c}{\textbf{Standard Deviation} ($\bm{\sigma}$)} \\
        \cline{2-5}
        & \bf GDE3 & \bf CO-GDE3 & \bf GDE3 & \bf CO-GDE3 \\
        \hline
        $HA$ & 3.48e-01 & \textbf{3.51e-01} & 4.50e-03 & \textbf{2.10e-03} \\
        $SP$ & \textbf{8.20e-03} & 8.30e-03 & 7.34e-04 & \textbf{6.38e-04} \\
        $FE_{min}$ & 512 & \textbf{380} & 248.86 & \textbf{152.75} \\
        \hline
    \end{tabular}
\end{table}

It is observed from TABLE \ref{tbl:trolleystats} and TABLE \ref{tbl:slewstats} that on average, the $HA$ and $SP$ values of both GDE3 and CO-GDE3 are quite competitive, with CO-GDE3 registering minor improvement (+1.48\% for T-MOTOP and +0.86\% for S-MOTOP) compared to GDE3 in terms of $\mu(HA)$ and no improvement (0\% for T-MOTOP and -1.20\% for S-MOTOP) in terms of $\mu(SP)$. This is expected as the variation operators (scaling factor $F$ and crossover operator $CR$) of GDE3 primarily guide the exploitation and exploration of the search space, and the initialization strategy contributes very little in enhancing the amplitude of the results. However, observing the $\sigma(HA)$ and $\sigma(SP)$ values, it can be inferred that CO-GDE3 is significantly more consistent (+58.18\% and +53.33\% in terms of $\sigma(HA)$, and +36.34\% and +13.07\% in terms of $\sigma(SP)$, for T-MOTOP and S-MOTOP, respectively) at finding the good $\mathcal{PF}_{known}$ in terms of solution quality and spread, compared to GDE3. Also, in Fig. \ref{fig:hyperareaplots} and Fig. \ref{fig:spacingplots}, CO-GDE3 has higher Q0, Q1, Q2, Q3 and Q4 of $HA$ for both the MOTOPs, lower Q2 and Q4 of $SP$ for T-MOTOP, and lower Q4 of $SP$ for S-MOTOP. This indicates that the proposed optimizer generates a superior quality of solutions for a higher percentage of simulation runs. This consistency, in contrast to the conventional GDE3, is due to the fact that each time, CO-GDE3 can form an initial solution spanning the entire search space, more effectively. Consequently, there is always a higher probability of having good solutions in the initial population of CO-GDE3, compared to the randomly generated initial population of GDE3. As the variation operators of the GDE3 employ information within the population to alter the search space through succeeding generations, CO-GDE3 can improve the frequency of attaining superior results.

The highest impact of the proposed initialization strategy is reflected in the values of the $FE_{min}$ metric. According to TABLE \ref{tbl:trolleystats}, for T-MOTOP, both $\mu(FE_{min})$ and $\sigma(FE_{min})$ of CO-GDE3 reports improvements of +22.97\% and +17.35\%, respectively. Similarly in TABLE \ref{tbl:slewstats}, for S-MOTOP, both $\mu(FE_{min})$ and $\sigma(FE_{min})$ of CO-GDE3 reports improvements of +25.78\% and +38.62\%, respectively. Since CO-GDE3 has a higher density of potentially good solutions than GDE3 in the initial population, it can find the Pareto optimal solutions in the objective space more quickly. This results in less number of function evaluations required for multiple runs, as demonstrated in Fig. \ref{fig:convergenceplots}. Hence, through the simulations, the CO-GDE3 emerges as the optimizer with a higher rank than the conventional GDE3.

\subsubsection{Multi-objective trajectory planning using the proposed planner}
\label{subsubsec:trajectoryresults}

\begin{figure}[tp]
    \centering
    \includegraphics[width=\columnwidth]{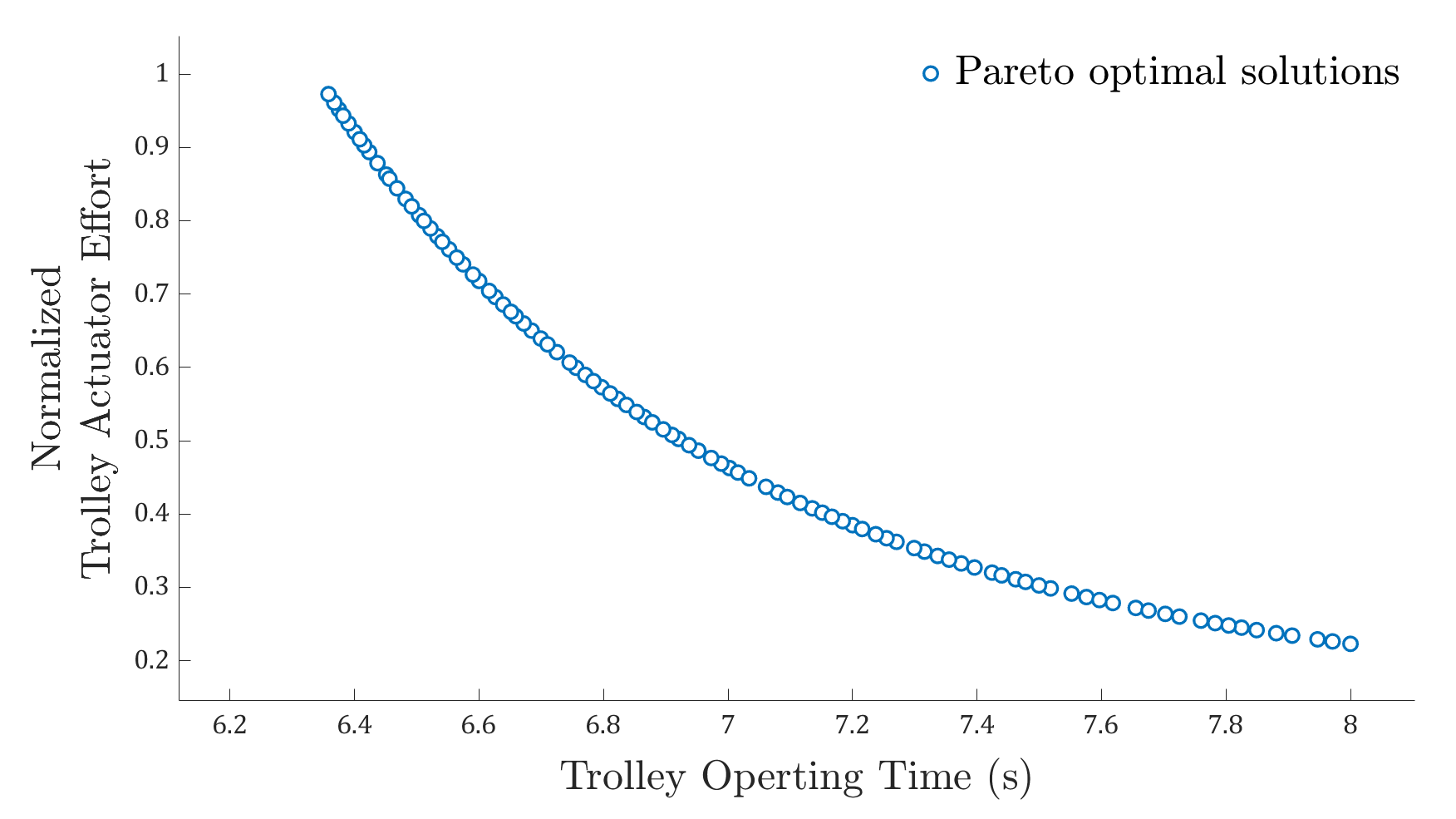}
    \caption{Pareto front obtained by CO-GDE3 optimizer for the T-MOTOP in TABLE \ref{tbl:operations} subjected to constraints in TABLE \ref{tbl:craneparameters}.}
    \label{fig:trolleyCOGDE3}
\end{figure}

\begin{figure}[tp]
    \centering
    \includegraphics[width=\columnwidth]{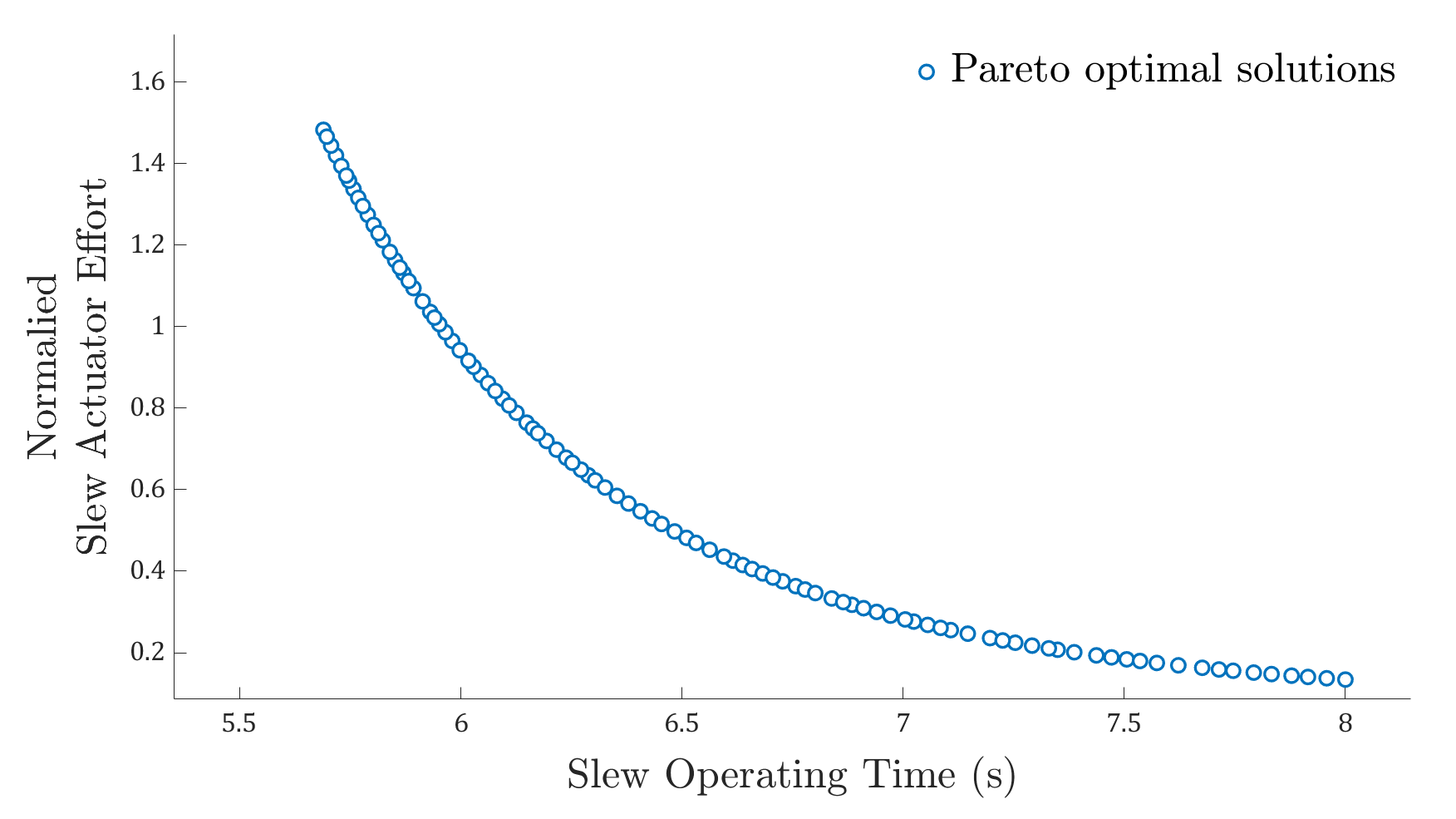}
    \caption{Pareto front obtained by CO-GDE3 optimizer for the S-MOTOP in TABLE \ref{tbl:operations} subjected to constraints in TABLE \ref{tbl:craneparameters}.}
    \label{fig:slewCOGDE3}
\end{figure}

For the T-MOTOP and the S-MOTOP, the best values of operating time and actuator effort satisfying all the constraints, obtained by the CO-GDE3 optimizer are illustrated in Fig. \ref{fig:trolleyCOGDE3} and Fig. \ref{fig:slewCOGDE3}, respectively. The scatter plots denote the $\mathcal{PF}_{known}$ the optimizer has converged to. The bounds of the Pareto fronts, ($f_{1min}$,$f_{2max}$) and ($f_{1max}$,$f_{2min}$), in the feasible objective space, for the two MOTOPs are listed in TABLE \ref{tbl:optimalsCOGDE3}. The maximum values of the mean linear fuzzy membership function $\Lambda_{max}$ for those fronts, and the corresponding optimal solutions selected via AFMF-based decision-maker, are also included in the aforementioned table. The boundary objective values together represent the feasible objective space. It is evident that for the trolley operation, the rate of change of required actuator effort (operating energy) w.r.t. operating time is low compared to that of the slew operation. So to obtain the benefit of higher productivity in terms of lower operating time, the slew actuating effort has to increase more than the trolley actuating effort.

\begin{table}[!t]
    \centering
    \caption{Optimal solutions provided by CO-GDE3 optimizer for the two MOTOPs in TABLE \ref{tbl:operations} subjected to constraints in TABLE \ref{tbl:craneparameters}.}
    \label{tbl:optimalsCOGDE3}
    \begin{tabular}{lcp{0.3\columnwidth}c}
        \hline \\[-3pt]
        \bf Problem & \bf Parameter & \bf Meaning & \bf Value \\[3pt]
        \hline \\[-3pt]
        \multirow{6}{*}{T-MOTOP} & ($f_{1min}$,$f_{2max}$) & (min time, max effort) & (6.36s, 0.97) \\[3pt]
        & ($f_{1max}$,$f_{2min}$) & (max time, min effort) & (8s, 0.22) \\[3pt]
        & $\Lambda_{max}$  & max mean fuzzy membership & 0.64 \\[3pt]
        & ($f_{1opt},f_{2opt}$) & (optimal time, optimal effort) & (7s, 0.46) \\[3pt]
        \hline \\[-3pt]
        \multirow{6}{*}{S-MOTOP} & ($f_{1min}$,$f_{2max}$) & (min time, max effort) & (5.69s, 1.48) \\[3pt]
        & ($f_{1max}$,$f_{2min}$) & (max time, min effort) & (8s, 0.13) \\[3pt]
        & $\Lambda_{max}$  & max mean fuzzy membership & 0.69 \\[3pt]
        & ($f_{1opt},f_{2opt}$) & (optimal time, optimal effort) & (6.51s, 0.48) \\[3pt]
        \hline 
    \end{tabular}
\end{table}

\begin{figure}[!t]
    \centering
    \subfloat[Trolley trajectories\label{subfig:trolleytrajectories}]{%
        \centering
        \includegraphics[width=\columnwidth]{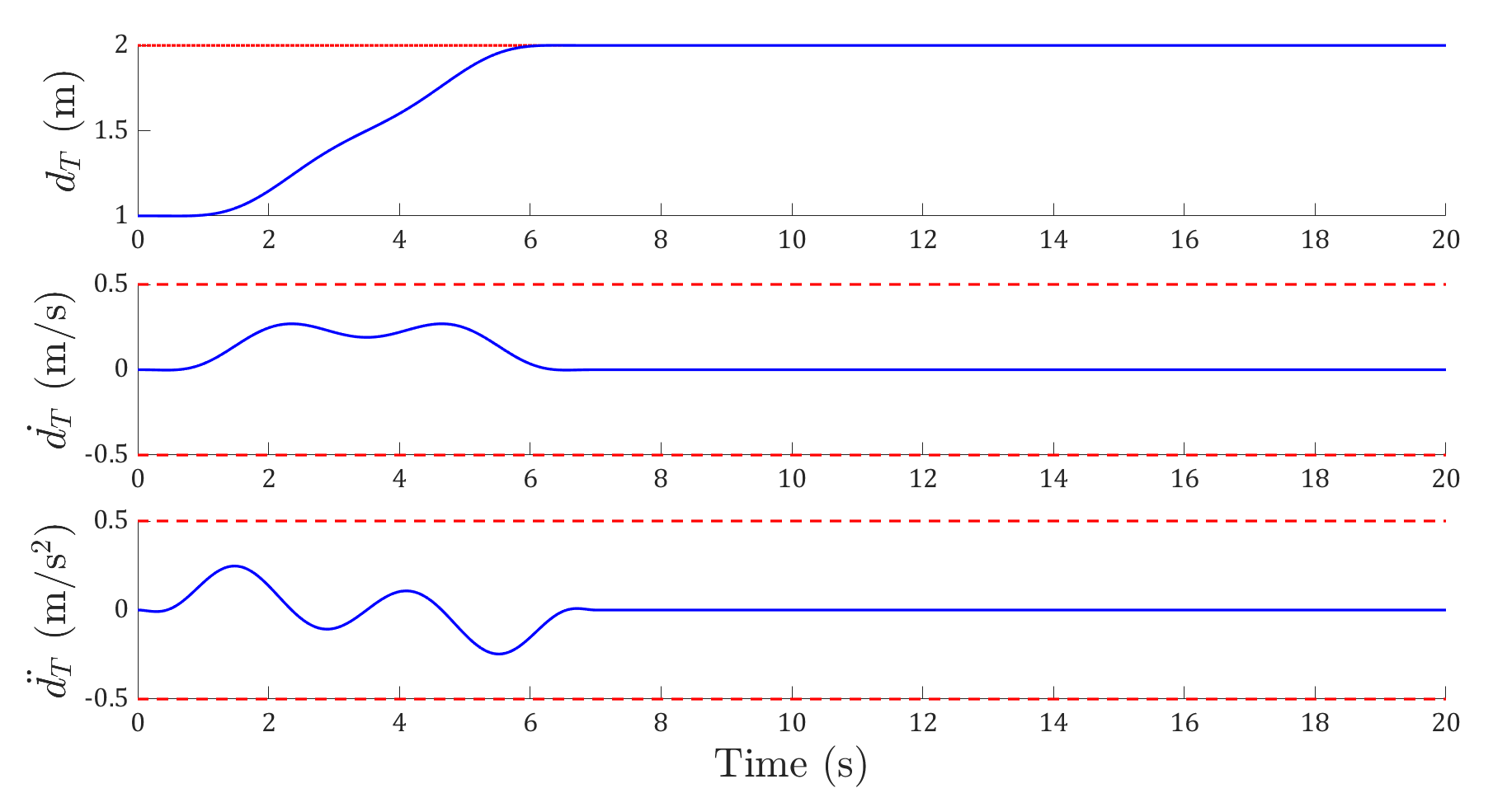}}
    \\
    \subfloat[Hook and payload swing trajectories\label{subfig:trolleyswingtrajectories}]{%
        \centering
        \includegraphics[width=\columnwidth]{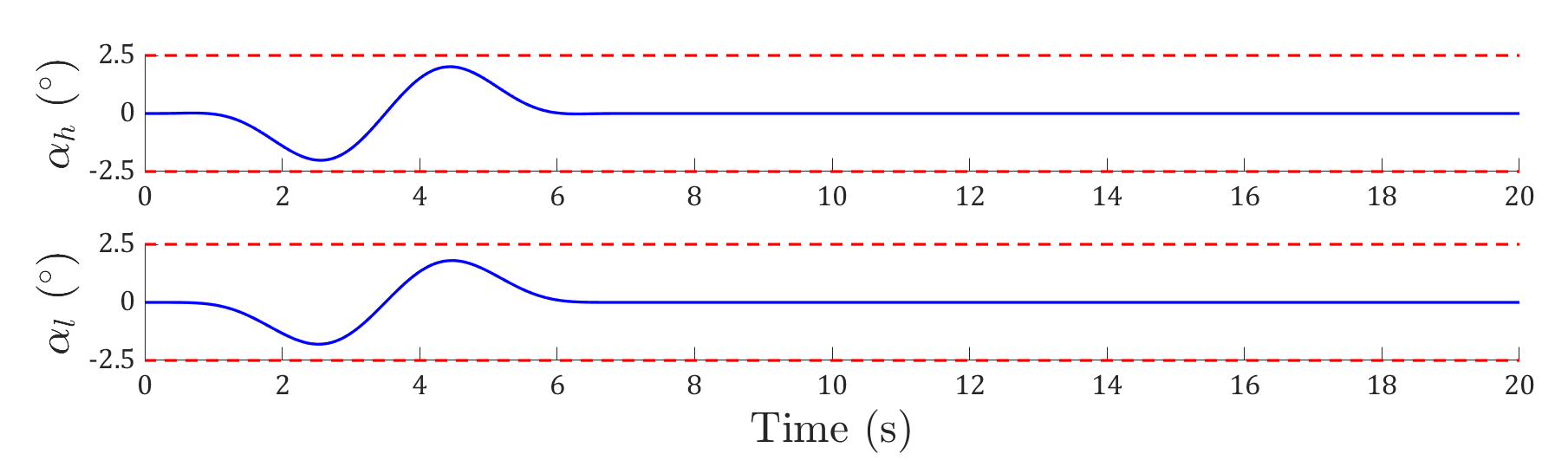}}
    \caption{Optimal trajectories obtained by the proposed anti-swing trajectory planner for (a) the trolley and (b) the radial swings of the hook and the payload during the trolley operation. (Red dotted line denotes the desired end position of the trolley, and red dashed lines indicate the safety constraint limits on the swing angles of the hook and the payload.)}
    \label{fig:trolleytrajectories}
\end{figure}

Based on the best fit of optimal time and energy for the T-MOTOP obtained in TABLE \ref{tbl:optimalsCOGDE3}, the optimal trajectories of the state variables during the trolley operation are plotted in Fig. \ref{fig:trolleytrajectories}. The velocity and acceleration of the trolley motion are kept well within their limits for the whole operation. The trolley velocity is always unidirectional (positive in this case as the trolley moves away from the tower along the jib) throughout the operation, confirming that the trolley does not alter the direction of motion during actuation. Moreover, the radial swing angles of the hook and the payload are also bounded by the safe ranges specified by the allowed payload deflection. The swing angles and their velocities and accelerations converge to zero as the trolley approaches the desired end position, omitting any residual swing.

\begin{figure}[!t]
    \centering
    \subfloat[Slew trajectories\label{subfig:slewtrajectories}]{%
        \centering
        \includegraphics[width=\columnwidth]{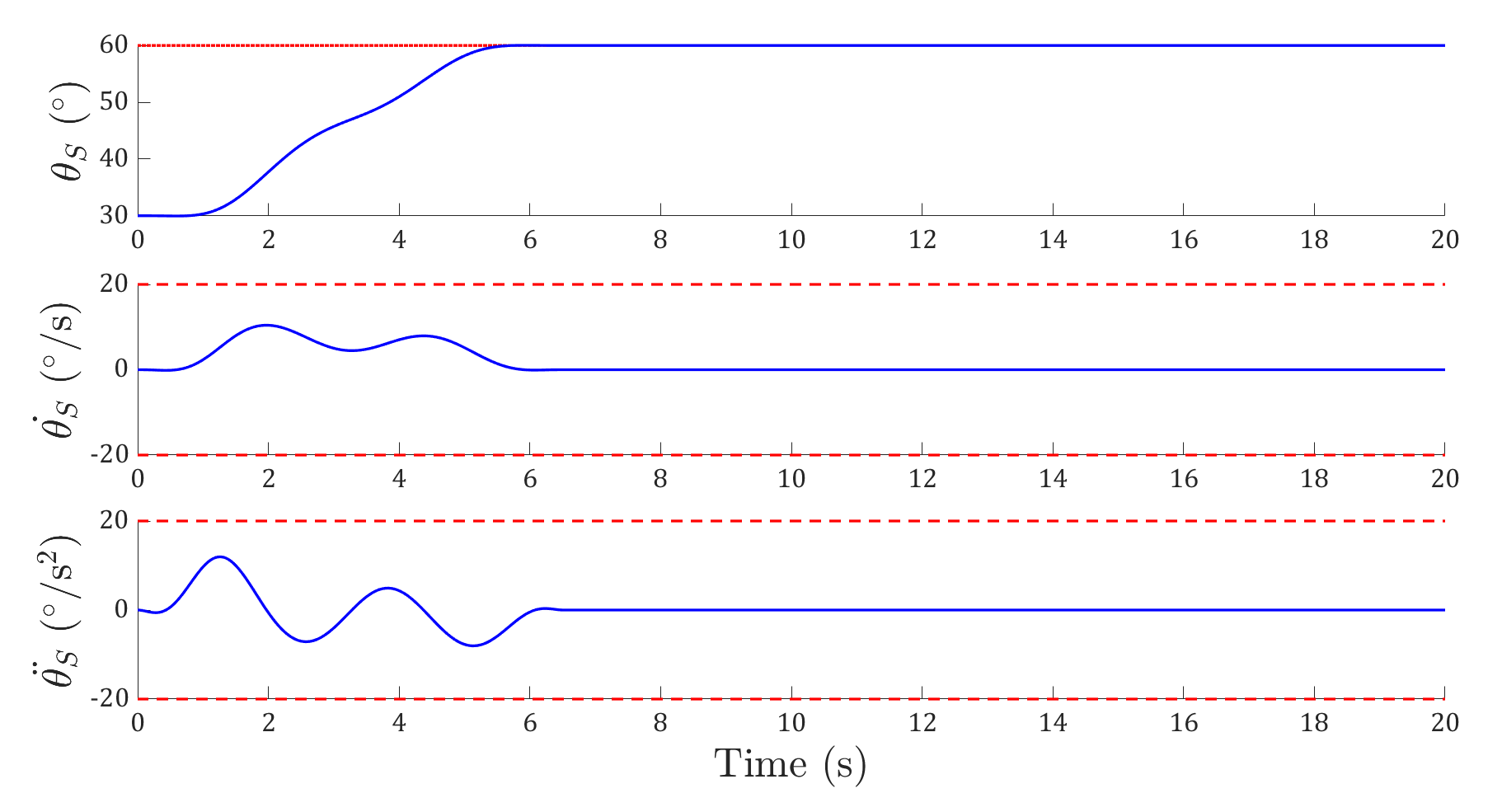}}
    \\
    \subfloat[Hook and payload swing trajectories\label{subfig:slewswingtrajectories}]{%
        \centering
        \includegraphics[width=\columnwidth]{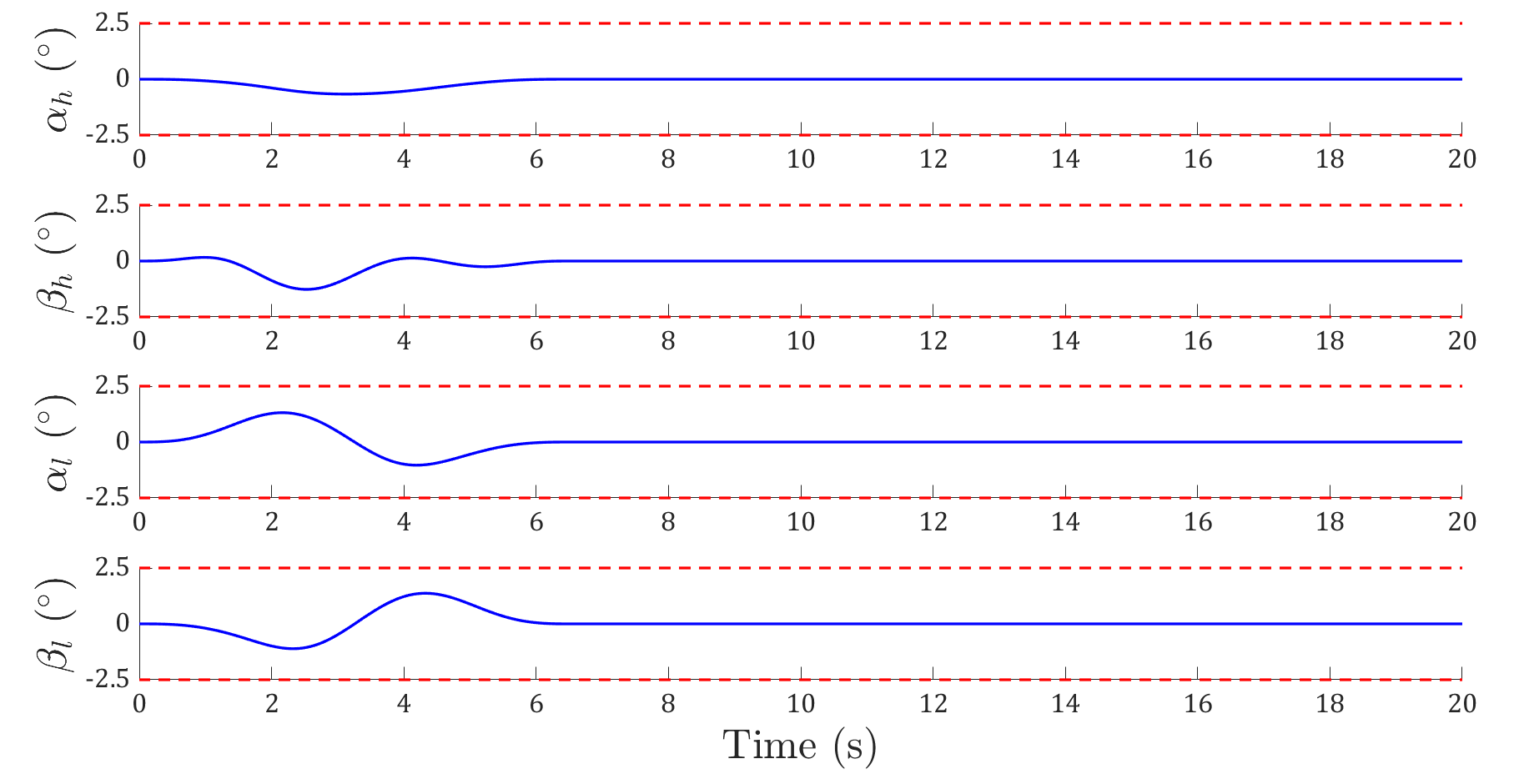}}
    \caption{Optimal trajectories obtained by the proposed anti-swing trajectory planner for (a) the jib and (b) the radial swings of the hook and the payload during the slew operation. (Red dotted line denotes the desired end position of the jib, and the red dashed lines indicate the safety constraint limits on the swing angles of the hook and the payload.)}
    \label{fig:slewtrajectories}
\end{figure}

For the S-MOTOP, the optimal trajectories of the jib are provided in Fig. \ref{fig:slewtrajectories}, where the slew motion can be seen to abide by the constraints of the tower crane slew actuator. Slew velocity is also unidirectional with positive values as the operation is performed in the counter-clockwise direction about the $z$ axis (from the perspective of the crane top view). In addition to the radial swing angles, the tangential swings are also shown to be effectively suppressed between the corresponding constraints. Similar to the scenario of trolley operation, there is no indication of residual swing from the hook or the payload.

It is to be noted, due to the absence of any previous work on multi-objective optimal trajectory planning of double-pendulum tower cranes, no additional comparisons could be performed for the optimal trajectories. Moreover, as an offline computation of optimal reference trajectories, the influence of any external disturbances or parametric uncertainties of the double-pendulum tower crane system is beyond the scope of the proposed method. Due to a lack of availability of necessary equipment, the hardware implementation of the planner with a conventional tracking controller was infeasible, and further validation of the trajectory planning method through physical experiments is aimed as a future work. To deal with internal as well as external uncertainties, a robust closed-loop tracking controller is currently under development to compensate for the system behaviour due to the disturbance inputs. Nevertheless, through numerical simulation studies, the proposed anti-swing double-pendulum tower crane trajectory planner has been shown to produce effective and efficient multi-objective reference trajectories. The optimal planner can constrain both the hook and the payload swings during trolley/jib positioning and suppress any residual swing. Such optimal reference trajectories can be used by the tower crane controller to track planned optimal collision-free lifting paths.

\section{Conclusion and Future Work}
\label{sec:conclusion}

In autonomous construction environments, tower cranes produce unactuated spherical double-pendulum hook-payload motion during lifting, when the hook and the payload are of comparable mass or the rig-cable is of non-negligible length. It is crucial yet difficult to suppress the corresponding swing angles during operation and at the end, due to the high nonlinear coupling between all the state variables. The present research work proposes possibly the \textit{first} multi-objective anti-swing trajectory planner for autonomous double-pendulum tower cranes. It can plan time-energy optimal trajectories, considering all the transient constraints imposed on the operations owing to the crane's mechanical limits and path safety. The planner deals with the complex dynamics of the double-swing behaviour of the cable-hook-payload unit during trolley/jib positioning by carefully considering the flat outputs of the double-pendulum system. The subsequently formulated constrained MOTOPs are solved by CO-GDE3, which is an improvement on GDE3 with a novel population initialization method. The proposed method, which exploits the higher probabilities of opposite solutions being closer to the Pareto optimal points, makes CO-GDE3 faster and more consistent than its conventional counterpart. Simulated trajectory results demonstrate that the planner can produce multi-objective optimal trajectories, keeping all the state variables within their respective limits, and restricting the hook-payload swings. In future work, a robust feedback tracking controller, which executes the operations accurately along the optimal trajectories generated by the proposed planner, is to be developed to tackle the parametric and non-parametric uncertainties of the system.

\bibliographystyle{IEEEtran}
\bibliography{preprint}

\end{document}